\documentclass[letterpaper]{article} 
\usepackage{aaai25}  
\usepackage{times}  
\usepackage{helvet}  
\usepackage{courier}  
\usepackage[hyphens]{url}  
\usepackage{graphicx} 
\usepackage{natbib}  
\usepackage{caption} 
\usepackage{amssymb}
\usepackage{amsmath}
\usepackage{multirow}
\usepackage{pifont}
\usepackage{subfigure}
\usepackage{overpic}
\usepackage{makecell}

\usepackage{color}
\usepackage{xcolor}
\usepackage[linesnumbered,ruled]{algorithm2e}
\usepackage{float}

\usepackage{newfloat}
\usepackage{listings}
\DeclareCaptionStyle{ruled}{labelfont=normalfont,labelsep=colon,strut=off} 
\floatstyle{ruled}
\newfloat{listing}{tb}{lst}{}
\floatname{listing}{Listing}
\title{\includegraphics[height=10pt]{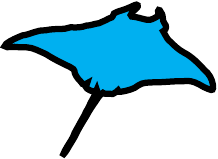} Manta: Enhancing Mamba for Few-Shot Action Recognition of \\ Long Sub-Sequence}
\author{
Wenbo~Huang\textsuperscript{\rm 1},
Jinghui~Zhang\textsuperscript{\rm 1}\thanks{Corresponding author.},
Guang~Li\textsuperscript{\rm 2},
Lei~Zhang\textsuperscript{\rm 3},
Shuoyuan~Wang\textsuperscript{\rm 4},
Fang~Dong\textsuperscript{\rm 1},
Jiahui~Jin\textsuperscript{\rm 1},
Takahiro~Ogawa\textsuperscript{\rm 2}, 
Miki~Haseyama\textsuperscript{\rm 2}
}
\affiliations{
\textsuperscript{\rm 1}Southeast University, Nanjing 211189, Jiangsu, China\\
\textsuperscript{\rm 2}Hokkaido University, Sapporo 060-0808, Hokkaido, Japan\\
\textsuperscript{\rm 3}Nanjing Normal University, Nanjing 210023, Jiangsu, China\\
\textsuperscript{\rm 4}Southern University of Science and Technology, Shenzhen 518055, Guangdong, China

wenbohuang1002@outlook.com, \{jhzhang,~jjin,~fdong\}@seu.edu.cn, \\
\{guang,~ogawa,~mhaseyama\}@lmd.ist.hokudai.ac.jp, leizhang@njnu.edu.cn, claytonwang0205@gmail.com
}

\usepackage{bibentry}

\begin{document}

\maketitle

\begin{abstract}
	In few-shot action recognition~(FSAR), long sub-sequences of video naturally express entire actions more effectively. However, the high computational complexity of mainstream Transformer-based methods limits their application. Recent Mamba demonstrates efficiency in modeling long sequences, but directly applying Mamba to FSAR overlooks the importance of local feature modeling and alignment. Moreover, long sub-sequences within the same class accumulate intra-class variance, which adversely impacts FSAR performance. To solve these challenges, we propose a \underline{\textbf{M}}atryoshka M\underline{\textbf{A}}mba and Co\underline{\textbf{N}}tras\underline{\textbf{T}}ive Le\underline{\textbf{A}}rning framework~(\textbf{Manta}). Firstly, the Matryoshka Mamba introduces multiple Inner Modules to enhance local feature representation, rather than directly modeling global features. An Outer Module captures dependencies of timeline between these local features for implicit temporal alignment. Secondly, a hybrid contrastive learning paradigm, combining both supervised and unsupervised methods, is designed to mitigate the negative effects of intra-class variance accumulation. The Matryoshka Mamba and the hybrid contrastive learning paradigm operate in two parallel branches within Manta, enhancing Mamba for FSAR of long sub-sequence. Manta achieves new state-of-the-art performance on prominent benchmarks, including SSv2, Kinetics, UCF101, and HMDB51. Extensive empirical studies prove that Manta significantly improves FSAR of long sub-sequence from multiple perspectives.
\end{abstract}

\section{Introduction}
\label{sec: introduction}
\indent Few-shot action recognition~(FSAR) addresses the labeling reliance in  data-driven training by classifying unseen actions from few video samples. This approach is widely used in real-world applications such as intelligent surveillance, video understanding, and health monitoring~\cite{liu2019exploring,reddy2022mumuqa,croitoru2021teachtext}. Long video sub-sequences intuitively offer advantages in expressing the entire process of an action, like “Diving cliff”, due to richer contextual information. In contrast, shorter sub-sequences may only cover partial actions, such as “Falling”, “Running”, and “Swimming”. Despite this, research on the usage of long sub-sequences in FSAR remains unexplored.\\
\begin{figure}[t]
	\centering
	\includegraphics[width=0.47\textwidth]{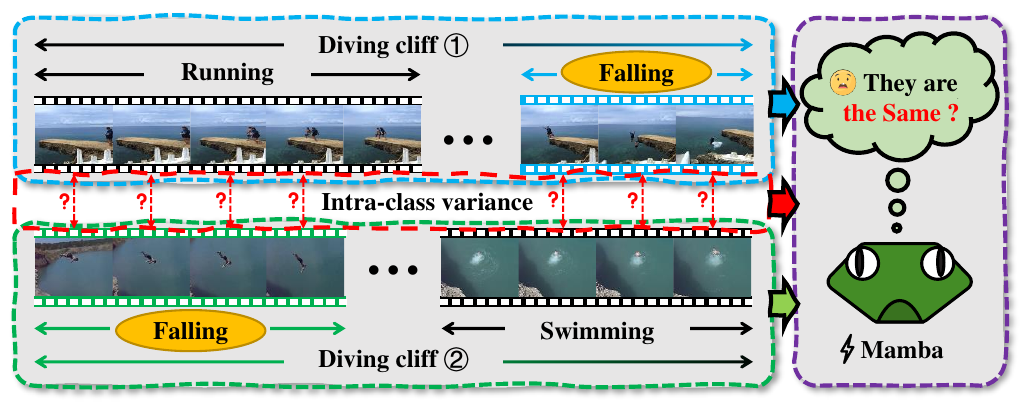} 
	\caption{In two long sub-sequence examples of “Diving cliff”, significant local features (highlighted as “Falling”) occupy only small portions of the examples and are located at different points in the timeline. Additionally, the frame pairs from these examples exhibit large discrepancies in visual features. As the number of frames increases, intra-class variance gradually accumulates.}
	\label{fig: long}
\end{figure}
\indent Mainstream Transformer-based methods~\cite{vaswani2017attention} for FSAR are limited to processing short sub-sequences of around 8 frames due to their computational complexity. Recently, Mamba~\cite{gu2023mamba,dao2024trans} has been applied to various tasks for its efficient long-sequence modeling capabilities without adding heavy computational overhead. Leveraging state space models~(SSMs)~\cite{gu2023train}, Mamba not only eliminates the complex attention mechanisms of Transformers but also flexibly manages the propagation and discarding of contextual information. However, the emphasis on global feature modeling by data-driven training in Mamba is misaligned with the extremely limited sample availability in FSAR. \\
\indent Therefore, while applying Mamba to FSAR with long sub-sequences appears promising, it still faces two inherent challenges, as illustrated in \figurename~\ref{fig: long}. \emph{\textbf{Challenge 1: The absence of local feature modeling and alignment.}} Though inconspicuous, some local features are crucial for accurate recognition. In two examples of “Diving cliff”, the core local features associated with “Falling” constitute only a small portion of the long sub-sequence, with the majority being secondary features. Focusing on global feature by Mamba often overlooks these critical local features, leading to potential misclassification in FSAR. Additionally, the core “Falling” features in different samples are not aligned temporally, and the absence of temporal alignment in Mamba exacerbates this issue, significantly degrading performance. \emph{\textbf{Challenge 2: The intra-class variance accumulation of long sub-sequences.}} Influenced by factors such as shooting conditions or post-processing, frame pairs between different “Diving cliff” examples of long sub-sequences exhibit significant visual discrepancies, regardless of alignment. As the number of frames increases, intra-class variance gradually accumulates, making it more challenging to cluster samples of the same class and leading to possible misclassification.\\
\indent Metric-based meta-learning is the mainstream paradigm in FSAR for efficacy and simplicity. After feature extraction, it embeds support samples into class prototypes for calculating distances between query samples, performing classification. Previous works directly apply explicit temporal alignment between sub-sequences~\cite{cao2020few,xing2023revisiting}, inevitably ignoring local features. To improve this issue, recent works focus on the combination of global and local features, achieving satisfactory results~\cite{perrett2021temporal,wang2022hybrid,wang2023molo}. However, we observe that they all utilize Transformer for short sub-sequence and are limited by complex calculation. In addition, the accumulation of intra-class variance is not severe due to the short sub-sequences, which are constantly overlooked. So far, solutions to the above challenges are absent. \\
\indent Based on these observations, we propose the \underline{\textbf{M}}atryoshka M\underline{\textbf{A}}mba and Co\underline{\textbf{N}}tras\underline{\textbf{T}}ive Le\underline{\textbf{A}}rning framework~(\textbf{Manta}). Firstly, the Matryoshka Mamba employs multiple Inner Modules to enhance local features instead of directly modeling global feature. An Outer Module is designed for implicit temporal alignment by capturing dependencies of timeline between local features. Secondly, a hybrid contrastive learning paradigm, which simultaneously incorporates both supervised and unsupervised methods, is developed to mitigate the impact of intra-class variance accumulation. The Matryoshka Mamba and the hybrid contrastive learning paradigm operate in parallel branches within Manta to enhance Mamba for FSAR of long sub-sequence.\\
\indent To the best of our knowledge, Manta is the first work to apply long sub-sequences and Mamba in FSAR. Our key contributions are threefold.
\begin{itemize}
	\item We propose the Matryoshka Mamba for local feature modeling and alignment. The Inner Modules enhance local features from fragments of a long sub-sequence, while the Outer Module bidirectionally scans the entire sequence to perform implicit temporal alignment through fusion. The nesting of Inner Modules within the Outer Module makes the Matryoshka Mamba a more suitable model for FSAR of long sub-sequence. 
	\item We design a hybrid contrastive learning paradigm for FSAR of long sub-sequence. Supervised contrastive learning is applied to labeled support samples, while an unsupervised method is used for unlabeled query samples. Subsequently, all samples are considered in an unsupervised manner. This approach enhances sample clustering and mitigates the negative impact of intra-class variance accumulation.
	\item Extensive experiments reveal that Manta achieves new state-of-the-art~(SOTA) performance on several FSAR benchmarks, including SSv2, Kinetics, UCF101, and HMDB51. Further analysis highlights competitiveness of Manta, particularly for long sub-sequences.
\end{itemize}
\section{Related works}
\label{sec: related works}
\subsection{Few-shot Action Recognition}
\label{subsec: fs image classification} 
\indent The mainstream paradigm of FSAR is metric-based meta-learning with Transformer to temporal alignment. Among them, OTAM~\cite{cao2020few} employs dynamic time warping~(DTW) algorithm to calculate sub-sequence similarities, aligning query and support samples. Then temporal relation is further emphasized, representative works are ITANet~\cite{zhang2021learning}, T$^2$AN~\cite{li2022ta2n}, and  STRM~\cite{thatipelli2022spatio}. To emphasize local features, fine-grained modeling is applied by TRX~\cite{perrett2021temporal}, HyRSM~\cite{wang2022hybrid}, SloshNet~\cite{xing2023revisiting}, and SA-CT~\cite{zhang2023importance}. Besides, additional information is introduced into the model, such as depth~\cite{fu2020depth}, optical flow~\cite{wanyan2023active}, and motion information~\cite{wang2023molo,wu2022motion}. Although remarkable performance was achieved, the above works are almost all based on short sub-sequences due to the computational complexity of Transformer architecture. 
\subsection{Mamba Architecture} 
\label{subsec: fs action recognition}
\indent Recently, Mamba with SSM~\cite{gu2023mamba,dao2024trans} has gained more attention because of its promising performance in modeling long sequences. More works are quickly moved to applying Mamba on vision tasks. Specifically, ViM~\cite{zhu2024vision} shares a similar idea with ViT, integrating Mamba into the vision model. Inspired by ResNet~\cite{he2016deep}, VMamba~\cite{liu2024vmamba} constructs a hierarchical structure with Mamba. To improve efficiency, EfficientVMamba~\cite{pei2024efficientvmamba} is designed by imposing selective scan. However, they are all unable to align local features in FSAR. 
\subsection{Contrastive Learning}
\label{subsec: contrastive} 
\indent In recent years, contrastive learning~\cite{he2020momentum,chen2020simple} receives increasing attention for its promising ability to learn generic representation from unlabeled samples. Subsequently, supervised contrastive learning~\cite{khosla2020supervised} makes full use of labels, accurately finding the positive and negative samples. Several works~\cite{zheng2022few,gidaris2019boosting,su2020does} point out that contrastive learning can serve as an auxiliary loss in few-shot learning, effectively alleviating the negative impact from intra-class variance. Hence, contrastive learning has the potential to solve the challenge of intra-class variance accumulation in FSAR.
\begin{figure*}[t]
	\centering
	\includegraphics[width=1\textwidth]{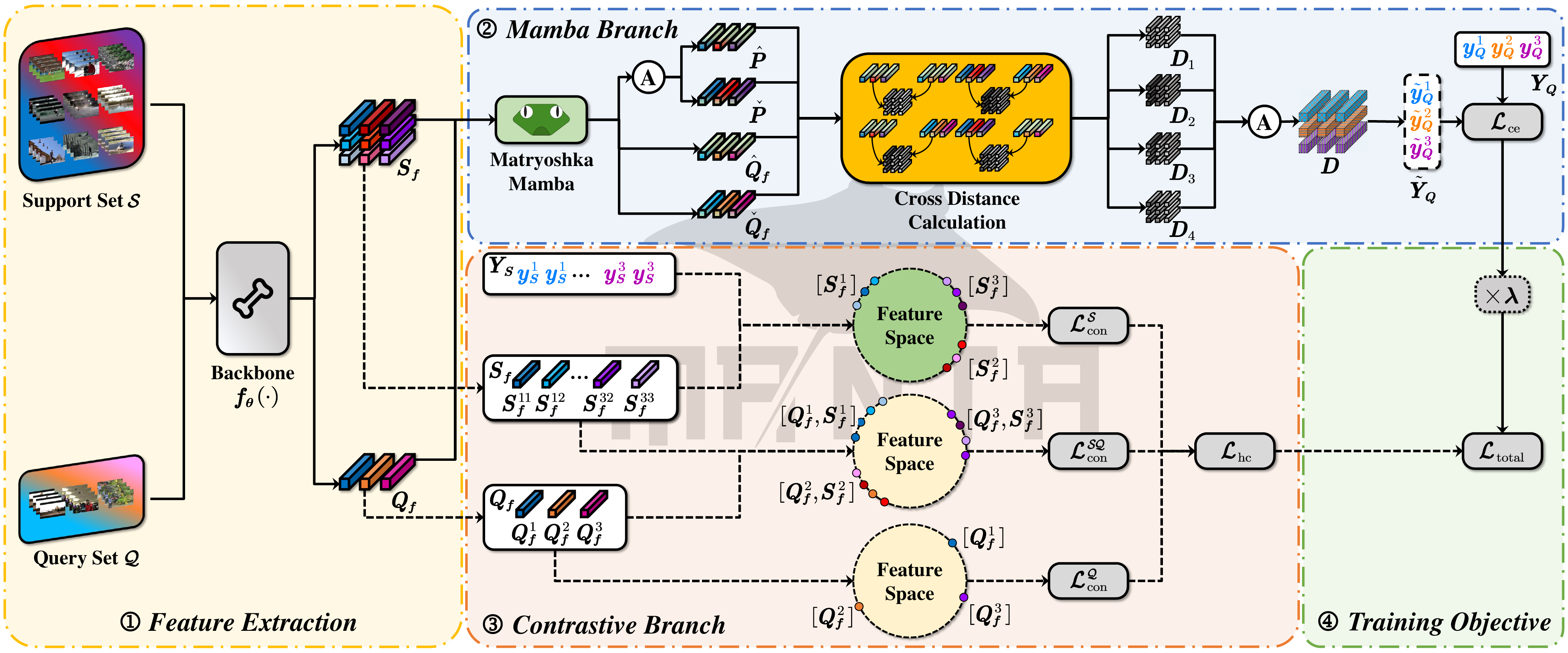} 
	\caption{The overall architecture of the Matryoshka Mamba and Contrastive Learning framework~(Manta) with four parts. To be specific, \normalsize{\textcircled{\scriptsize{\textbf{1}}}} Feature Extraction with backbone extracts features from query and support. \normalsize{\textcircled{\scriptsize{\textbf{2}}}} Mamba Branch with Matryoshka Mamba can emphasize local features and execute temporal alignment. \normalsize{\textcircled{\scriptsize{\textbf{3}}}} Contrastive Branch alleviates the accumulation of intra-class variance by hybrid contrastive learning. \normalsize{\textcircled{\scriptsize{\textbf{4}}}} Training Objective $\mathcal{L}_{\text{total}} $ is the loss combination of cross-entropy loss $\mathcal{L}_{\text{ce}} $ from \normalsize{\textcircled{\scriptsize{\textbf{2}}}} Mamba Branch and contrastive loss $\mathcal{L}_{\text{hc}} $ from \normalsize{\textcircled{\scriptsize{\textbf{3}}}} Contrastive Branch. Notion \normalsize{\textcircled{\scriptsize{\textbf{A}}}} means averaging calculation.}
	\label{fig: overall arch}
\end{figure*}
\section{Methodology}
\subsection{Problem Definition}
\label{subsec: problem def}
\indent According to previous works~\cite{cao2020few,perrett2021temporal}, the dataset is divided into three non-overlapping parts including training set $\mathcal{D}_\text{train}$, validation set $\mathcal{D}_\text{val}$, and test set $\mathcal{D}_\text{test}$~($\mathcal{D}_\text{train}\cap \mathcal{D}_\text{val}\cap \mathcal{D}_\text{test}=\varnothing $). In each part, classifying unlabeled samples from query set $\mathcal{Q}$ into one class of support set $\mathcal{S}$~($\mathcal{S}\cap \mathcal{Q}=\varnothing $) is the goal of FSAR. There is at least one labeled sample in each class of $\mathcal{S}$. In episodic training, a mass of few-shot tasks are randomly selected from $\mathcal{D}_\text{train}$. The $N$-way $K$-shot setting means that $\mathcal{S}$ in each task has $N$ classes and $K$ samples in each class. 
\subsection{Overall Architecture}
\label{subsec: over arch}
\indent Figure~\ref{fig: overall arch} is an overview of Manta under 3-way 3-shot setting. A backbone is applied for feature extraction. In the Mamba branch, Matryoshka Mamba can enhance local feature modeling and alignment under various scales. Cross-entropy loss $\mathcal{L}_{\text{ce}} $ can be calculated by the distance between query and prototypes. In the contrastive branch, supervised and unsupervised paradigms work simultaneously, achieving hybrid contrastive learning loss $\mathcal{L}_{\text{hc}} $. The training objective $\mathcal{L}_{\text{total}} $ is the weighted combination of $\mathcal{L}_{\text{ce}} $ and $\mathcal{L}_{\text{hc}} $.
\begin{figure}[ht]
	\centering
	\includegraphics[width=0.33\textwidth]{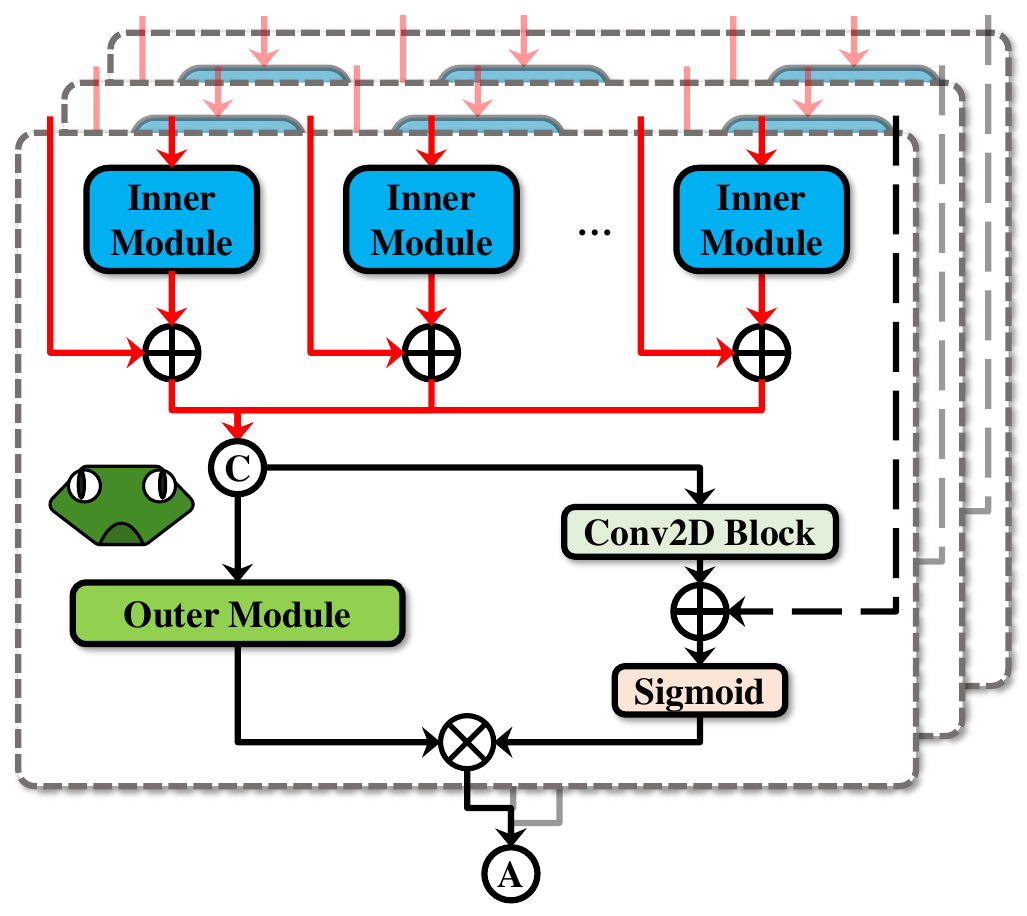} 
	\caption{The structure of Matryoshka Mamba, \normalsize{\textcircled{\small{\textbf{+}}}}, \normalsize{\textcircled{\small{\textbf{$\times$}}}}, and \normalsize{\textcircled{\scriptsize{\textbf{C}}}} indicate element-wise addition, multiplication and concatenate operation. Conv2D Block has three 2D convolutions and a batch normalization layer. Red indicates local features while feature itself is dotted line.}
	\label{fig: mamba}
\end{figure}
\subsection{Feature Extraction}
\label{subsec: feature}
\indent Sub-sequences with $F$ frames are uniformly sampled from a video each time. The $k^\text{th}$ ($k = 1, \cdots, K$) support sample $S^{ck}$ in the $c^\text{th}$ ($c = 1, \cdots, N$) class of the support set $\mathcal{S}$ and the randomly selected query sample $Q^{r}$ ($r \in \mathbb{Z}^+$) from the query set $\mathcal{Q}$ are defined as follows:
\begin{equation}
	\begin{aligned}
		S^{ck}&=\left[ s_{1}^{ck},\dots ,s_{F}^{ck} \right] \in \mathbb{R} ^{F\times C\times H\times W}, \\
		Q^{r}&=\left[ q_{1}^{r},\dots ,q_{F}^{r} \right] \in \mathbb{R} ^{F\times C\times H\times W}.
	\end{aligned}
	\label{eq1}
\end{equation}
Notions applied are $F$~(frames), $C$~(channels), $H$~(height), and $W$~(width), respectively. $S^{ck}$ and $Q^{r}$ are sent into backbone $f_{\theta}\left( \cdot \right) : \mathbb{R} ^{C\times H\times W}\mapsto \mathbb{R} ^D $ for $D$-dimensional vectors $S_{f}^{ck}, Q_{f}^{r} \in \mathbb{R} ^{F\times D}$:
\begin{equation}
	\begin{aligned}
		S_{f}^{ck}&=\left[ f_{\theta}\left( s_{1}^{ck} \right) ,\dots ,f_{\theta}\left( s_{F}^{ck} \right) \right],
		\\
		Q_{f}^{r}&=\left[ f_{\theta}\left( q_{1}^{r} \right) ,\dots ,f_{\theta}\left( q_{F}^{r} \right) \right].
	\end{aligned}
	\label{eq2}
\end{equation}
\subsection{Mamba Branch}
\label{subsec: Mamba}
\indent The structure of Matryoshka Mamba is shown in \figurename~\ref{fig: mamba}. Multiple Inner Modules for local feature modeling are nested within an Outer Module for alignment, designed with Mamba-2 for high efficiency. Other models including Mamba-1 can also be utilized. The above structure is designed under a fixed scale. Building on this foundation, we extend the single scale to multiple scales for more comprehensive local feature modeling and alignment. The set $\mathcal{O}$ is defined as a hyper-parameter of multi-scale. The cardinality $\left| \mathcal{O} \right|$ denotes number of scales, while an element $o$ ($o \in \mathcal{O}$, $F \mid o$, $o < F$, $o = 2^\alpha$, and $\alpha \in \mathbb{Z}^+$) represents the frame count at this scale. For simplicity, we will use the subscript $o$ to indicate an arbitrary scale.
\subsubsection{Space State Models.}
\begin{figure}[t]
	\centering
	\includegraphics[width=0.32\textwidth]{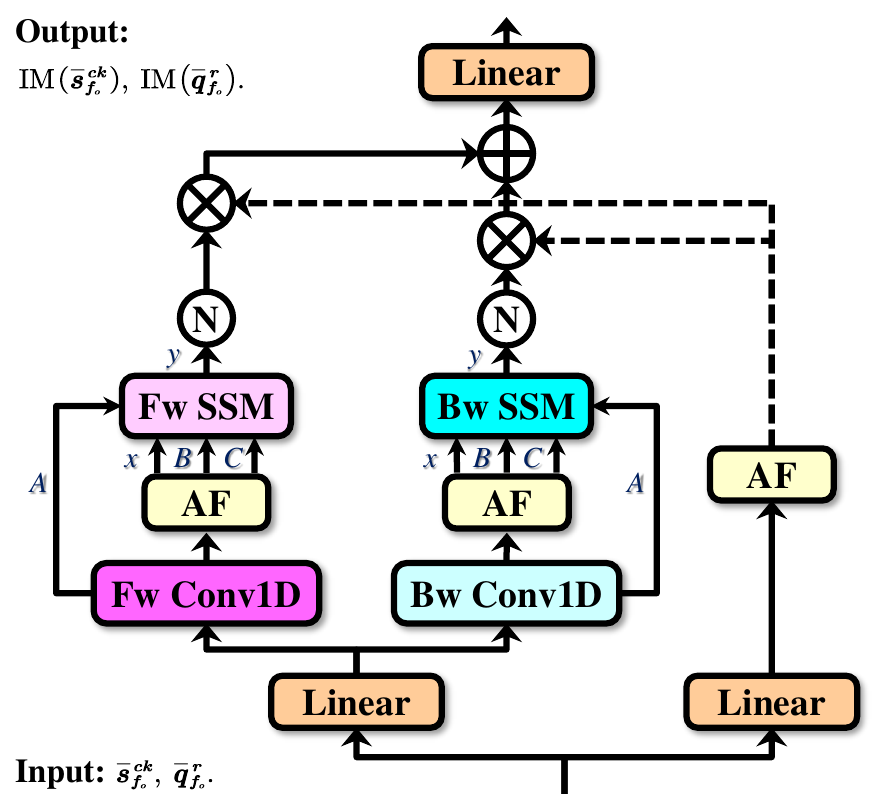} 
	\caption{The structure of Inner Module based on Mamba-2, where \normalsize{\textcircled{\scriptsize{\textbf{N}}}}, Fw, Bw, AF, and SSM refers to normalization, forward, backward, activation function, and state space model.}
	\label{fig: umamba}
\end{figure}
\indent SSM in Mamba can effectively handle local feature modeling and alignment by flexibly propagating or discarding contextual information. It transforms input $x(t)\in \mathbb{R}^L$ to output $y(t)\in \mathbb{R}^L$ through hidden states $h(t)\in \mathbb{R}^H$. Linear ordinary differential equations are used for SSM description.
\begin{equation}
	\begin{aligned}
		h^{\prime}\left( t \right) =Ah\left( t \right) +Bx\left( t \right) ,
		y\left( t \right) =Ch\left( t \right) .
	\end{aligned}
	\label{eq3}
\end{equation}
$h^{\prime}\left( t \right)$ is the derivative of $h\left( t \right)$. $A\in \mathbb{R}^{H\times H}$ is the state transition matrix while $B, C\in \mathbb{R}^H$ are projection parameters. 
\subsubsection{Inner Module.}
\indent The feature of a long sub-sequence is divided into non-overlapping fragments, with Inner Modules enhancing local features from each fragment. As shown in \figurename~\ref{fig: umamba}, the Inner Modules consist of two sub-branches, $\mathrm{IM}_{\text{Fw}}\left( \cdot \right)$ and $\mathrm{IM}_{\text{Bw}}\left( \cdot \right)$, which do not share parameters due to the differences in forward and backward local feature modeling. $\bar{s}_{f_o}^{ck}, \bar{q}_{f_o}^{r}\in \mathbb{R}^{o\times D}$ represent local feature tensors with length $L=D$ and hidden state $H=o$. The output is $\mathrm{IM}\left( \cdot \right) =\mathrm{Linear}\left[ \mathrm{IM}_{\mathrm{Fw}}\left( \cdot \right) \oplus \mathrm{IM}_{\mathrm{Bw}}\left( \cdot \right) \right] \in \mathbb{R} ^{F\times o}$. Concatenating~($\mathrm{C}\left[ \dots ,\cdot ,\dots \right]$) each enhanced local feature and then adding with input, $\tilde{S}_{fo}^{ck}, \tilde{Q}_{fo}^{r} \in \mathbb{R} ^{F\times D}$ can also be seen as tensors with length $L=F$ and hidden state $H=D$.
\begin{equation}
	\begin{aligned}
		\tilde{S}_{f_o}^{ck}&=\mathrm{C}\left[ \dots ,\mathrm{IM}\left( \bar{s}_{f_o}^{ck} \right) \oplus \bar{s}_{f_o}^{ck},\dots \right] ,
		\\
		\tilde{Q}_{f_o}^{r}&=\mathrm{C}\left[ \dots ,\mathrm{IM}\left( \bar{q}_{f_o}^{r} \right) \oplus \bar{q}_{f_o}^{r},\dots \right] .
	\end{aligned}
	\label{eq4}
\end{equation}
\subsubsection{Outer Module.}
\begin{figure}[t]
	\centering
	\includegraphics[width=0.36\textwidth]{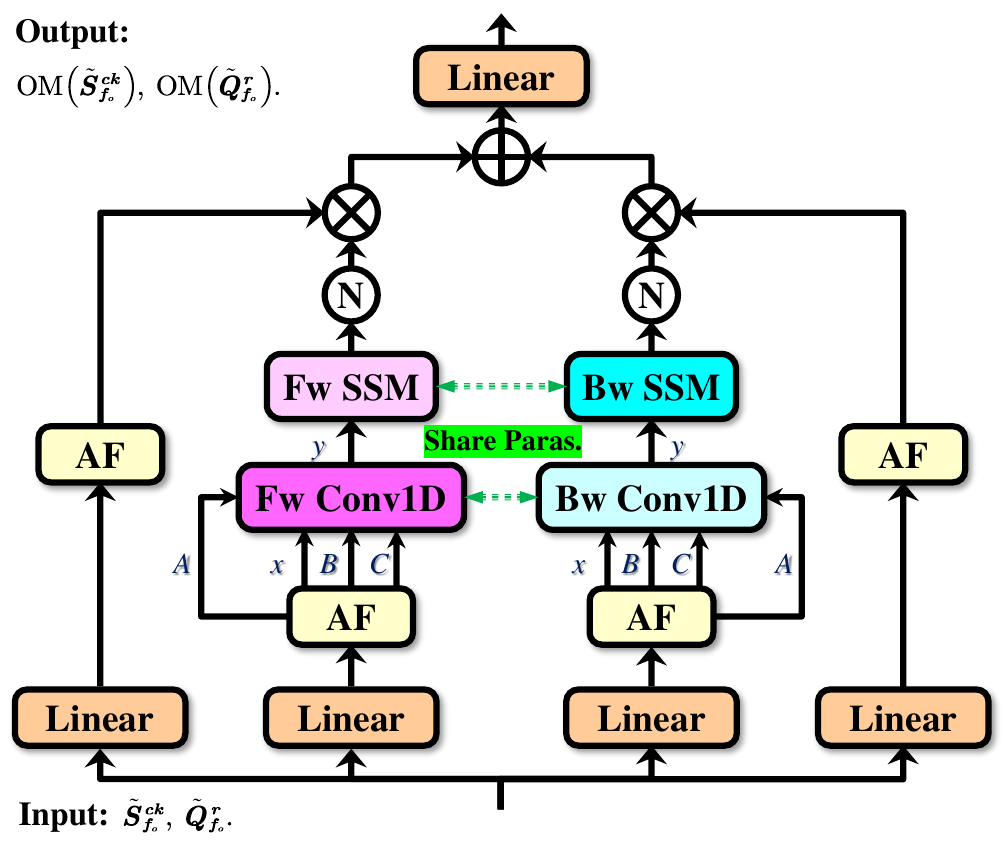} 
	\caption{The bidirectional structure of Outer Module based on Mamba-2, decomposing the input at first. Two sub-branches share parameters.}
	\label{fig: cmamba}
\end{figure}
\indent As illustrated in \figurename~\ref{fig: cmamba}, two sub-branches $\mathrm{OM}_{\text{Fw}}\left( \cdot \right), \mathrm{OM}_{\text{Bw}}\left( \cdot \right) $ with shared parameters are employed for bidirectional scanning since the forward and backward alignments are identical, capturing temporal dependencies for implicit alignment. After fusing them, the output is $\mathrm{OM}\left( \cdot \right)=\mathrm{Linear}\left[\mathrm{OM}_{\text{Fw}}\left( \cdot \right) \oplus \mathrm{OM}_{\text{Bw}}\left( \cdot \right)\right] \in \mathbb{R} ^{F\times D}$. Inspired by C3-STISR~\cite{zhao2022c3}, we design learnable weights $w_{o}^{S}, w_{o}^{Q}$ for weighted averaging of various scales. Learnable weights are calculated from the input of Outer Module and feature itself. We define the Conv2D Block as $\mathrm{CB(\cdot)}$. The above calculation is written as
\begin{equation}
	\begin{aligned}
		w_{o}^{S}&=\mathrm{Sigmoid}\left[ \mathrm{CB}\left( \tilde{S}_{f_o}^{ck} \right) \oplus S_{f}^{ck} \right] ,
		\\
		w_{o}^{Q}&=\mathrm{Sigmoid}\left[ \mathrm{CB}\left( \tilde{Q}_{f_o}^{r} \right) \oplus Q_{f}^{r} \right] .
	\end{aligned}
	\label{eq5}
\end{equation} 
Through combining with outputs of Outer Module, arbitrarily scaled $\mathring{S}_{f_o}^{ck}, \mathring{Q}_{f_o}^{r} \in \mathbb{R} ^{F\times D}$ are obtained, referring to
\begin{equation}
	\begin{aligned}
		\mathring{S}_{f_o}^{ck}=w_{o}^{S}\otimes \mathrm{OM}\left( \tilde{S}_{f_o}^{ck} \right), \;
		\mathring{Q}_{f_o}^{r}=w_{o}^{Q}\otimes \mathrm{OM}\left( \tilde{Q}_{f_o}^{r} \right).
	\end{aligned}
	\label{eq6}
\end{equation} 
The final outputs $\hat{S}_{f}^{ck}, \hat{Q}_{f}^{r} \in \mathbb{R} ^{F\times D}$ of Matryoshka Mamba are averaged from all scales as
\begin{equation}
	\begin{aligned}
		\hat{S}_{f}^{ck}=\frac{1}{\left| \mathcal{O} \right|}\sum_{o\in \mathcal{O}}{\mathring{S}_{f_o}^{ck}}, \;
		\hat{Q}_f^{r}=\frac{1}{\left| \mathcal{O} \right|}\sum_{o\in \mathcal{O}}{\mathring{Q}_{f_o}^{r}}.
	\end{aligned}
	\label{eq7}
\end{equation} 
\subsubsection{Prototype Construction.} The prototype of support $\hat{P}^{c}$ is constructed by the averaged paradigm as
\begin{equation}
	\begin{aligned}
		\hat{P}^c=\frac{1}{K}\sum_{k=1}^K{\hat{S}_{f}^{ck}.}
	\end{aligned}
	\label{eq8}
\end{equation}
\subsubsection{Cross Distance Calculation.} To further enhance temporal alignment, we apply cross-distance calculation. Specifically, $\check{P}^c, \check{Q}_{f}^{r}$ are the inversion of $\hat{P}^c, \hat{Q}_{f}^{r}$. Considering symmetrical alignment, a higher possibility of the same class is indicated by the smaller distances between tensors sharing the same superscript or the larger distances between tensors of various superscripts. Therefore, we take the reciprocal of those distances with various superscripts.
\begin{equation}
	\begin{aligned}
		D_1&=\left\| \hat{P}^c-\hat{Q}_{f}^{r} \right\| , \; D_2=\left\| \check{P}^c-\check{Q}_{f}^{r} \right\| ,
		\\
		D_3&=\left\| \hat{P}^c-\check{Q}_{f}^{r} \right\| ^{-1}, \; D_4=\left\| \check{P}^c-\hat{Q}_{f}^{r} \right\| ^{-1}.
	\end{aligned}
	\label{eq9}
\end{equation}
The distance $D$ between query and the $c^\text{th}$ support is the average of the above four distances. Therefore, the model can predict label $\tilde{y}_Q^j\in \tilde{Y}_Q$ of query as
\begin{equation}
	\begin{aligned}
		\tilde{y}_Q^j=\underset{c}{\mathrm{argmin}}\left( D \right), \; D= \frac{1}{4}\sum_{i=1}^4{D_i} .
	\end{aligned}
	\label{eq10}
\end{equation}
Cross-entropy loss $\mathcal{L}_\text{ce}$ is calculated from the predicted label $\tilde{y}_Q^j$ and the ground truth $y_Q^j\in Y_Q$.
\begin{equation}
	\begin{aligned}
		\mathcal{L} _{\text{ce}}=-\frac{1}{N}\sum_{j=1}^N{y_Q^j\log \left( \tilde{y}_Q^j \right)}.
	\end{aligned}
	\label{eq11}
\end{equation}
\subsection{Contrastive Branch} For alleviating the negative impact of intra-class variance accumulation, a hybrid contrastive learning paradigm with supervised and unsupervised methods is applied. Corresponding loss is formulated as
\begin{equation}
	\begin{aligned}
		\mathcal{L} _{\text{con}}=-\log \frac{e^{\text{sim}\left( z,z^\text{p} \right) /\tau}}{e^{\mathrm{sim}\left( z,z^\text{p} \right) /\tau}+\sum\nolimits_{r=1}^R{e^{\mathrm{sim}\left( z,z_{r}^{\text{n}} \right) /\tau}}},
	\end{aligned}
	\label{eq12}
\end{equation}
Here, $\mathrm{sim}\left( \cdot, \cdot \right)$ represents cosine similarity in the feature space, and $\tau$ denotes the temperature hyper-parameter. In the supervised paradigm applied to the support set, features with the same label are treated as positive samples $z^\text{p}$, while the other $R$ features with different labels are treated as negative samples $z^\text{n}$. In the unsupervised paradigm applied to the query set and all samples, there is no label guidance (the construction of $z^\text{p}$ and $z^\text{n}$ is detailed in the Supplementary Materials). The three contrastive losses including supervised $\mathcal{L} _{\text{con}}^{\mathcal{S}}$ and unsupervised $\mathcal{L} _{\text{con}}^{\mathcal{Q}}, \mathcal{L} _{\text{con}}^{\mathcal{SQ}}$ are combined to form the hybrid contrastive loss $\mathcal{L}_{\text{hc}}$.
\begin{equation}
	\begin{aligned}
		\mathcal{L} _{\text{hc}}=\mathcal{L} _{\text{con}}^{\mathcal{S}}+\mathcal{L} _{\text{con}}^{\mathcal{Q}}+\mathcal{L} _{\text{con}}^{\mathcal{SQ}}.
	\end{aligned}
	\label{eq13}
\end{equation}
\subsection{Training Objective}
\label{subsec: training}
\indent During the training stage, the above loss functions of two main branches supervise our Manta, the total loss $\mathcal{L}_\text{total}$ is
\begin{equation}
	\begin{aligned}
		\mathcal{L} _{\text{total}}=\lambda \mathcal{L} _{\text{ce}}+\mathcal{L} _{\text{hc}},
	\end{aligned}
	\label{eq14}
\end{equation}
where $\lambda$ means the weight factor of $\mathcal{L} _{\text{ce}}$. In summary, Matryoshka Mamba enhances local feature modeling and alignment, while the hybrid contrastive learning paradigm alleviates the negative impact of intra-class variance accumulation. These combined operations make Manta a more suitable framework for FSAR of long sub-sequence.
\section{Experiments}
\subsection{Experimental Configuration}
\begin{table*}[ht]
	\centering
	\small
	\begin{tabular}{l|r|cc|cc|cc|cc}
		\Xhline{1pt}
		\multirow{2}{*}{Methods} & \multicolumn{1}{c|}{\multirow{2}{*}{Pre-Backbone}} & \multicolumn{2}{c|}{SSv2} & \multicolumn{2}{c|}{Kinetics} & \multicolumn{2}{c|}{UCF101} & \multicolumn{2}{c}{HMDB51} \\ \cline{3-10} 
		& \multicolumn{1}{c|}{} & \multicolumn{1}{c|}{1-shot} & 5-shot & \multicolumn{1}{c|}{1-shot} & 5-shot & \multicolumn{1}{c|}{1-shot} & 5-shot & \multicolumn{1}{c|}{1-shot} & 5-shot \\ \hline
		STRM~\cite{thatipelli2022spatio} & ImageNet-RN50 & \multicolumn{1}{c|}{N/A} & 68.1 & \multicolumn{1}{c|}{N/A} & 86.7 & \multicolumn{1}{c|}{N/A} & 96.9 & \multicolumn{1}{c|}{N/A} & 76.3 \\
		SloshNet~\cite{xing2023revisiting} & ImageNet-RN50 & \multicolumn{1}{c|}{46.5} & 68.3 & \multicolumn{1}{c|}{N/A} & 87.0 & \multicolumn{1}{c|}{N/A} & 97.1 & \multicolumn{1}{c|}{N/A} & 77.5 \\
		SA-CT~\cite{zhang2023importance} & ImageNet-RN50 & \multicolumn{1}{c|}{48.9} & 69.1 & \multicolumn{1}{c|}{71.9} & 87.1 & \multicolumn{1}{c|}{85.4} & 96.3 & \multicolumn{1}{c|}{61.2} & 76.9 \\
		GCSM~\cite{yu2023multi} & ImageNet-RN50 & \multicolumn{1}{c|}{N/A} & N/A & \multicolumn{1}{c|}{74.2} & 88.2 & \multicolumn{1}{c|}{86.5} & 97.1 & \multicolumn{1}{c|}{61.3} & 79.3 \\
		GgHM~\cite{xing2023boosting} & ImageNet-RN50 & \multicolumn{1}{c|}{54.5} & 69.2 & \multicolumn{1}{c|}{74.9} & 87.4 & \multicolumn{1}{c|}{85.2} & 96.3 & \multicolumn{1}{c|}{61.2} & 76.9 \\
		STRM~\cite{thatipelli2022spatio} & ImageNet-ViT & \multicolumn{1}{c|}{N/A} & 70.2 & \multicolumn{1}{c|}{N/A} & 91.2 & \multicolumn{1}{c|}{N/A} & \underline{98.1} & \multicolumn{1}{c|}{N/A} & 81.3 \\
		SA-CT~\cite{zhang2023importance} & ImageNet-ViT & \multicolumn{1}{c|}{N/A} & \multicolumn{1}{c|}{66.3} & \multicolumn{1}{c|}{N/A} & \multicolumn{1}{c|}{91.2} & \multicolumn{1}{c|}{N/A} & \multicolumn{1}{c|}{98.0} & \multicolumn{1}{c|}{N/A} & \multicolumn{1}{c}{81.6} \\
		\cline{2-10}
		$^{\star}$TRX~\cite{perrett2021temporal} & ImageNet-RN50 & \multicolumn{1}{c|}{53.8} & 68.8 & \multicolumn{1}{c|}{74.9} & 85.9 & \multicolumn{1}{c|}{85.7} & 96.3 & \multicolumn{1}{c|}{63.5} & 75.8 \\
		$^{\star}$HyRSM~\cite{wang2022hybrid} & ImageNet-RN50 & \multicolumn{1}{c|}{54.1} & 68.7 & \multicolumn{1}{c|}{73.5} & 86.2 & \multicolumn{1}{c|}{83.6} & 94.6 & \multicolumn{1}{c|}{60.1} & 76.2 \\
		$^{\star}$MoLo~\cite{wang2023molo} & ImageNet-RN50 & \multicolumn{1}{c|}{56.6} & 70.7 & \multicolumn{1}{c|}{74.2} & 85.7 & \multicolumn{1}{c|}{86.2} & 95.4 & \multicolumn{1}{c|}{67.1} & 77.3 \\
		$^{\star}$TRX~\cite{perrett2021temporal} & ImageNet-ViT & \multicolumn{1}{c|}{57.2} & 71.4 & \multicolumn{1}{c|}{76.3} & 87.5 & \multicolumn{1}{c|}{\underline{88.9}} & 97.2 & \multicolumn{1}{c|}{66.9} & 78.8 \\
		$^{\star}$HyRSM~\cite{wang2022hybrid} & ImageNet-ViT & \multicolumn{1}{c|}{58.8} & 71.3 & \multicolumn{1}{c|}{76.8} & 92.3 & \multicolumn{1}{c|}{86.6} & 96.4 & \multicolumn{1}{c|}{69.6} & 82.2 \\
		$^{\star}$MoLo~\cite{wang2023molo} & ImageNet-ViT & \multicolumn{1}{c|}{61.1} & 71.7 & \multicolumn{1}{c|}{78.9} & \underline{95.8} & \multicolumn{1}{c|}{88.4} & 97.6 & \multicolumn{1}{c|}{\underline{71.3}} & 84.4 \\
		$^{\star}$TRX~\cite{perrett2021temporal} & ImageNet-VM & \multicolumn{1}{c|}{56.9} & 71.5 & \multicolumn{1}{c|}{76.2} & 87.2 & \multicolumn{1}{c|}{88.1} & 97.0 & \multicolumn{1}{c|}{66.7} & 78.5 \\
		$^{\star}$HyRSM~\cite{wang2022hybrid} & ImageNet-VM & \multicolumn{1}{c|}{58.6} & 71.7 & \multicolumn{1}{c|}{76.6} & 92.4 & \multicolumn{1}{c|}{86.8} & 96.5 & \multicolumn{1}{c|}{70.2} & 82.6 \\
		$^{\star}$MoLo~\cite{wang2023molo} & ImageNet-VM & \multicolumn{1}{c|}{\underline{61.3}} & \underline{72.1} & \multicolumn{1}{c|}{\underline{79.4}} & 95.6 & \multicolumn{1}{c|}{88.2} & 97.4 & \multicolumn{1}{c|}{71.1} & \underline{84.5} \\
		\hline
		AmeFu-Net~\cite{fu2020depth} & ImageNet-RN50 & \multicolumn{1}{c|}{N/A} & N/A & \multicolumn{1}{c|}{74.1} & 86.8 & \multicolumn{1}{c|}{85.1} & 95.5 & \multicolumn{1}{c|}{60.2} & 75.5 \\
		MTFAN~\cite{wu2022motion} & ImageNet-RN50 & \multicolumn{1}{c|}{45.7} & 60.4 & \multicolumn{1}{c|}{74.6} & 87.4 & \multicolumn{1}{c|}{84.8} & 95.1 & \multicolumn{1}{c|}{59.0} & 74.6 \\
		AMFAR~\cite{wanyan2023active} & ImageNet-RN50 & \multicolumn{1}{c|}{\underline{61.7}} & \underline{79.5} & \multicolumn{1}{c|}{\underline{80.1}} & \underline{92.6} & \multicolumn{1}{c|}{\underline{91.2}} & \underline{99.0} & \multicolumn{1}{c|}{\underline{73.9}} & \underline{87.8} \\
		\cline{2-10}
		$^{\star}$Lite-MKD~\cite{liu2023lite} & ImageNet-RN50 & \multicolumn{1}{c|}{55.7} & 69.9 & \multicolumn{1}{c|}{75.0} & 87.5 & \multicolumn{1}{c|}{85.3} & 96.8 & \multicolumn{1}{c|}{66.9} & 74.7 \\
		$^{\star}$Lite-MKD~\cite{liu2023lite} & ImageNet-ViT & \multicolumn{1}{c|}{59.1} & 73.6 & \multicolumn{1}{c|}{78.8} & 90.6 & \multicolumn{1}{c|}{89.6} & 98.4 & \multicolumn{1}{c|}{71.1} & 77.4 \\
		$^{\star}$Lite-MKD~\cite{liu2023lite} & ImageNet-VM & \multicolumn{1}{c|}{59.3} & 73.8 & \multicolumn{1}{c|}{78.5} & 90.8 & \multicolumn{1}{c|}{90.1} & 98.6 & \multicolumn{1}{c|}{71.5} & 77.2 \\
		\hline
		Manta~(Ours) & ImageNet-RN50 & \multicolumn{1}{c|}{\textbf{63.4}} & \textbf{87.4} & \multicolumn{1}{c|}{\textbf{82.4}} & \textbf{94.2} & \multicolumn{1}{c|}{\textbf{95.9}} & \textbf{99.2} & \multicolumn{1}{c|}{\textbf{86.8}} & \textbf{96.4} \\ 
		Manta~(Ours) & ImageNet-ViT & \multicolumn{1}{c|}{\textbf{66.2}} & \textbf{89.3} & \multicolumn{1}{c|}{\textbf{84.2}} & \textbf{96.3} & \multicolumn{1}{c|}{\textbf{97.2}} & \textbf{99.5} & \multicolumn{1}{c|}{\textbf{88.9}} & \textbf{96.8} \\
		Manta~(Ours) & ImageNet-VM & \multicolumn{1}{c|}{\textbf{66.1}} & \textbf{89.1} & \multicolumn{1}{c|}{\textbf{84.4}} & \textbf{96.2} & \multicolumn{1}{c|}{\textbf{96.9}} & \textbf{99.4} & \multicolumn{1}{c|}{\textbf{89.1}} & \textbf{96.6} \\
		\Xhline{1pt}
	\end{tabular}
	\caption{Comparison~($\uparrow$ Acc.~\%) on ResNet-50~(ImageNet-RN50), ViT-B~(ImageNet-ViT), and VMamba-B~(ImageNet-VM). \textbf{Bold texts} denotes the global best results. From top to bottom, the whole table is divided into three parts including RGB-based, multimodal, and our Manta. In the first two parts, “$\star$” represents our implementation with the same setting. “N/A” indicates not available in the corresponding publication. \underline{Underline texts} serve as the local best results.}
	\label{tab: acc com}
\end{table*}
\subsubsection{Data Processing.} 
\indent Widely used benchmark datasets such as temporal-related SSv2~\cite{goyal2017something}, spatial-related Kinetics~\cite{carreira2017quo}, UCF101~\cite{soomro2012ucf101}, and HMDB51~\cite{kuehne2011hmdb} are selected for proving the effectiveness of Manta. The sampling intervals are set to each 1 frame when decoding videos. According to the most common data split~\cite{zhu2018compound,cao2020few,zhang2020few}, all datasets are divided into $\mathcal{D}_\text{train}$, $\mathcal{D}_\text{val}$, and $\mathcal{D}_\text{test}$~($\mathcal{D}_\text{train}\cap \mathcal{D}_\text{val}\cap \mathcal{D}_\text{test}=\varnothing $). Tasks of FSAR aim to classify query samples $Q^r$ into corresponding classes of support set $\mathcal{S}$.\\
\indent On the basis of TSN~\cite{wang2016temporal}, frames are resized to $3\times256\times256$. $F$ frames per video are sequentially sampled each time. For simulating sub-sequences with various lengths, $F$~($F\in \left[ 8,128 \right] $, $F\mid 2$) can be adjusted according to actual situation. In the regular setup, $F\geqslant 16$ can be seen as long sub-sequences. Data is augmented with $3\times224\times224$ random crops and horizontal flipping during training, while only the center crop is employed for testing. Due to SSv2 having many actions with horizontal direction such as “Pushing S from right to left”\footnote{“S” denotes “something”.}, horizontal flipping is absent in this dataset~\cite{cao2020few}. 
\subsubsection{Implementation Details and Evaluation Metrics.} 
\indent We adopt two standard few-shot settings including 5-way 1-shot and 5-shot to conduct experiments. For a comprehensive comparison, ResNet-50~\cite{he2016deep}, ViT-B~\cite{dosovitskiy2020image}, and VMamba-B~\cite{liu2024vmamba} initialized with pre-trained weights on ImageNet~\cite{deng2009imagenet} are served as the backbone. Features extracted are 2048-dimensional vectors~($D=2048$). \\ 
\indent Except for the larger SSv2 which requires 75,000 tasks training, other datasets utilize 10,000 tasks. An SGD optimizer with an initial learning rate of $10^{-3}$ is applied for training. The $\mathcal{D}_\text{val}$ determines hyper-parameters including multi-scale~($\mathcal{O}=\left\{ 1,2,4 \right\}$), temperature~($\tau=0.07$) and weight factor of loss~($\lambda=4$). During the stage of testing, average accuracy across 10,000 random tasks of the testing set is reported. Most experiments are completed on a server with two 32GB NVIDIA Tesla V100 PCIe GPUs.
\subsection{Comparison with Various Methods}
\label{subsec: result}
For a fair comparison with recent methods, we set the sub-sequence length as $F=8$ and employ various backbones in this part. The average accuracy~($\uparrow$ higher indicates better) is demonstrated in \tablename~\ref{tab: acc com}. Experiments on long sub-sequence are conducted in subsequent parts. 
\subsubsection{ResNet-50 Methods.} 
\indent Using the SSv2 dataset under 1-shot as an example, we observe that our Manta with ResNet-50 backbone improves the current SOTA method AMFAR from 61.7\% to 63.4\%. It is worth mentioning that AMFAR is a multimodal method with much heavier computational complexity than Manta. A similar improvement can also be observed in other datasets under various few-shot settings. 
\subsubsection{ViT-B Methods.} 
\indent In FSAR, ViT-B has fewer applications than ResNet-50. Methods using ViT-B tend to outperform because of its larger model capacity. For instance, in the 5-shot Kinetics dataset, the previous SOTA performance for RGB-based methods was achieved by MoLo. Following a similar trend with ResNet-50, Manta demonstrates superior performance, even surpassing the multimodal AMFAR.
\subsubsection{VMamba-B Methods.} 
\indent As an emerging model, VMamba gains increasing attention for efficient feature extraction ability. Therefore, further comparison based on VMamba-B is also conducted in FSAR. Its performance is better than ResNet-50 because of a larger capacity. Our Manta with VMamba-B also achieves a remarkable improvement on performance, performing better than other VMamba-B based and even multimodal methods.
\subsection{Essential Components and Factors}
\label{subsec: ablation}
\indent All 1-shot experiments in this part are all trained and tested with ResNet-50 of long sub-sequences ($F=16$).  
\subsubsection{Key Components.}
\begin{table}[t]
	\centering
	\small
	\begin{tabular}{c|c|c|cc|cc}
		\Xhline{1pt}
		\multirow{2}{*}{IM} & \multirow{2}{*}{OM} & \multirow{2}{*}{$\mathcal{L} _{\text{hc}}$} & \multicolumn{2}{c|}{SSv2} & \multicolumn{2}{c}{Kinetics} \\ \cline{4-7} 
		&  &  & \multicolumn{1}{c|}{1-shot} & 5-shot & \multicolumn{1}{c|}{1-shot} & 5-shot \\ \hline
		\ding{55} & \ding{55} & \ding{55} & \multicolumn{1}{c|}{46.3} & 64.5 & \multicolumn{1}{c|}{70.9} & 86.5 \\ \hline
		\ding{51} & \ding{55} & \ding{55} & \multicolumn{1}{c|}{55.5} & 69.3 & \multicolumn{1}{c|}{75.3} & 88.2 \\
		\ding{55} & \ding{51} & \ding{55} & \multicolumn{1}{c|}{55.3} & 69.4 & \multicolumn{1}{c|}{75.0} & 88.0 \\
		\ding{55} & \ding{55} & \ding{51} & \multicolumn{1}{c|}{48.0} & 64.9 & \multicolumn{1}{c|}{72.1} & 87.4 \\ \hline
		\ding{51} & \ding{51} & \ding{55} & \multicolumn{1}{c|}{63.8} & 88.0 & \multicolumn{1}{c|}{82.8} & 94.6 \\
		\ding{51} & \ding{55} & \ding{51} & \multicolumn{1}{c|}{62.3} & 87.1 & \multicolumn{1}{c|}{82.3} & 93.8 \\
		\ding{55} & \ding{51} & \ding{51} & \multicolumn{1}{c|}{61.9} & 86.7 & \multicolumn{1}{c|}{81.9} & 93.4 \\ \hline
		\ding{51} & \ding{51} & \ding{51} & \multicolumn{1}{c|}{\textbf{64.7}} & \textbf{88.7} & \multicolumn{1}{c|}{\textbf{84.1}} & \textbf{96.2} \\ 
		\Xhline{1pt}
	\end{tabular}
	\caption{Comparison~($\uparrow$ Acc.~\%) of key components. }
	\label{tab: key}
\end{table}
\indent To verify the effect of key components, we split Manta into Inner Module~(IM), Outer Module~(OM), and hybrid contrastive learning loss~($\mathcal{L} _{\text{hc}}$) for testing. As indicated in \tablename~\ref{tab: key}, we have the following observation. Each key component improves the performance of the model. Therefore, applying the whole Manta brings the largest improvement by emphasizing local features, executing alignment, and reducing the negative impact of intra-class variance accumulation.
\subsubsection{Multi-Scale Design.}
\begin{table}[t]
	\centering
	\small
	\begin{tabular}{l|cc|cc}
		\Xhline{1pt}
		\multirow{2}{*}{Multi-Scale} & \multicolumn{2}{c|}{SSv2} & \multicolumn{2}{c}{Kinetics} \\ \cline{2-5} 
		& \multicolumn{1}{c|}{1-shot} & 5-shot & \multicolumn{1}{c|}{1-shot} & 5-shot \\ \hline
		$\mathcal{O}=\left\{ 1 \right\}$ & \multicolumn{1}{c|}{63.3} & 87.2 & \multicolumn{1}{c|}{81.6} & 92.9 \\
		$\mathcal{O}=\left\{ 2 \right\}$ & \multicolumn{1}{c|}{63.4} & 87.0 & \multicolumn{1}{c|}{81.3} & 93.2 \\
		$\mathcal{O}=\left\{ 4 \right\}$ & \multicolumn{1}{c|}{63.2} & 87.3 & \multicolumn{1}{c|}{81.4} & 93.1 \\
		$\mathcal{O}=\left\{ 8 \right\}$ & \multicolumn{1}{c|}{63.0} & 86.9 & \multicolumn{1}{c|}{81.1} & 92.7 \\
		\hline
		$\mathcal{O}=\left\{ 1, 2 \right\}$ & \multicolumn{1}{c|}{63.8} & 88.2 & \multicolumn{1}{c|}{82.8} & 94.6 \\
		$\mathcal{O}=\left\{ 1, 4 \right\}$ & \multicolumn{1}{c|}{64.1} & 87.9 & \multicolumn{1}{c|}{83.4} & 94.5 \\
		$\mathcal{O}=\left\{ 1, 8 \right\}$ & \multicolumn{1}{c|}{64.0} & 87.6 & \multicolumn{1}{c|}{82.8} & 94.3 \\
		$\mathcal{O}=\left\{ 2, 4 \right\}$ & \multicolumn{1}{c|}{63.9} & 88.4 & \multicolumn{1}{c|}{83.2} & 94.3 \\
		$\mathcal{O}=\left\{ 2, 8 \right\}$ & \multicolumn{1}{c|}{63.6} & 87.4 & \multicolumn{1}{c|}{82.7} & 94.1 \\
		$\mathcal{O}=\left\{ 4, 8 \right\}$ & \multicolumn{1}{c|}{63.6} & 87.3 & \multicolumn{1}{c|}{82.8} & 94.0 \\
		\hline
		$\mathcal{O}=\left\{ 1, 2, 4 \right\}$ & \multicolumn{1}{c|}{\textbf{64.7}} & \textbf{88.7} & \multicolumn{1}{c|}{\textbf{84.1}} & \textbf{96.2} \\
		$\mathcal{O}=\left\{ 1, 2, 8 \right\}$ & \multicolumn{1}{c|}{64.5} & 88.5 & \multicolumn{1}{c|}{83.8} & 96.0 \\
		$\mathcal{O}=\left\{ 2, 4, 8 \right\}$ & \multicolumn{1}{c|}{64.4} & 88.2 & \multicolumn{1}{c|}{83.6} & 95.8 \\
		\hline
		$\mathcal{O}=\left\{ 1, 2, 4, 8 \right\}$ & \multicolumn{1}{c|}{64.1} & 88.2 & \multicolumn{1}{c|}{83.4} & 95.6 \\
		\Xhline{1pt}
	\end{tabular}
	\caption{Comparison~($\uparrow$ Acc.~\%) of multi-scale design.}
	\label{tab: scale}
\end{table}
\indent In \tablename~\ref{tab: scale}, experiments with various hyper-parameter $\mathcal{O}$ are conducted for exploring how multi-scale design affect Manta. We find that emphasizing local features can improve performance. When multi-scale is introduced, the improvement will be further expanded, as $\left| \mathcal{O} \right|=3$ performs better than $\left| \mathcal{O} \right|=2$ while $\left| \mathcal{O} \right|=2$ defeats $\left| \mathcal{O} \right|=1$. However, $\left| \mathcal{O} \right|=4$ cannot further improve performance due to the redundant information. Therefor, we determine that $\mathcal{O}=\left\{ 1, 2, 4 \right\}$ is the $\mathcal{O}$ setting without redundancy, performing better than other settings.
\subsubsection{Temporal Alignment.}
\begin{figure}[ht]
	\centering
	\subfigure[SSv2]{\includegraphics[width=0.23\textwidth]{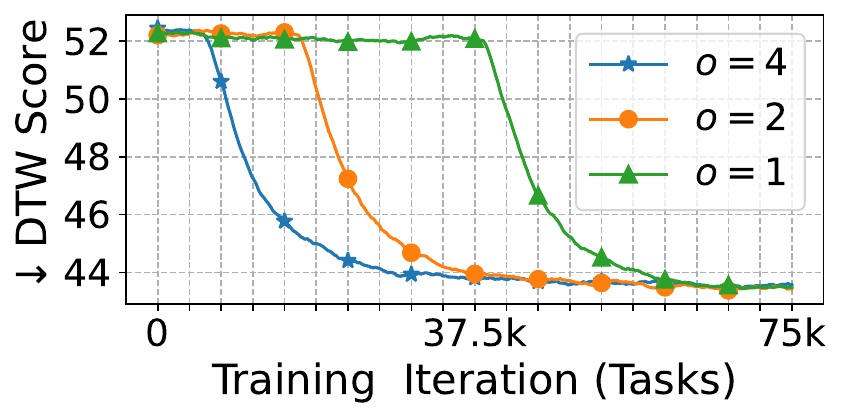}}
	\subfigure[Kinetics]{\includegraphics[width=0.23\textwidth]{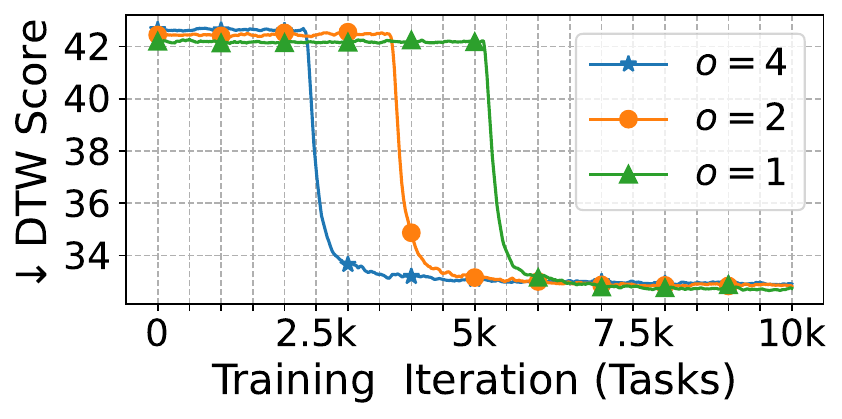}}
	\caption{Alignment~($\downarrow$ DTW scores) during training. $o=4$, $o=2$, and $o=1$ mean large, middle, and small scales.}
	\label{fig: dtw}
\end{figure}

\indent These experiments are designed to reveal the trend of alignment from large to small scales. According to OTAM~\cite{cao2020few}, we calculate DTW scores~($\downarrow$~lower indicates better) from the shortest aligned paths of different scales. As illustrated in \figurename~\ref{fig: dtw}, the DTW scores of two datasets all decline as training progresses, indicating that Manta learns temporal alignment gradually. In addition, the points of decline and convergence are sorted from early to late, proving that the alignment process of Manta is executed from coarse-grained to fine-grained.
\subsubsection{Lengthening Sub-Sequence.}
\begin{figure}[ht]
	\centering
	\subfigure[SSv2]{\includegraphics[width=0.23\textwidth]{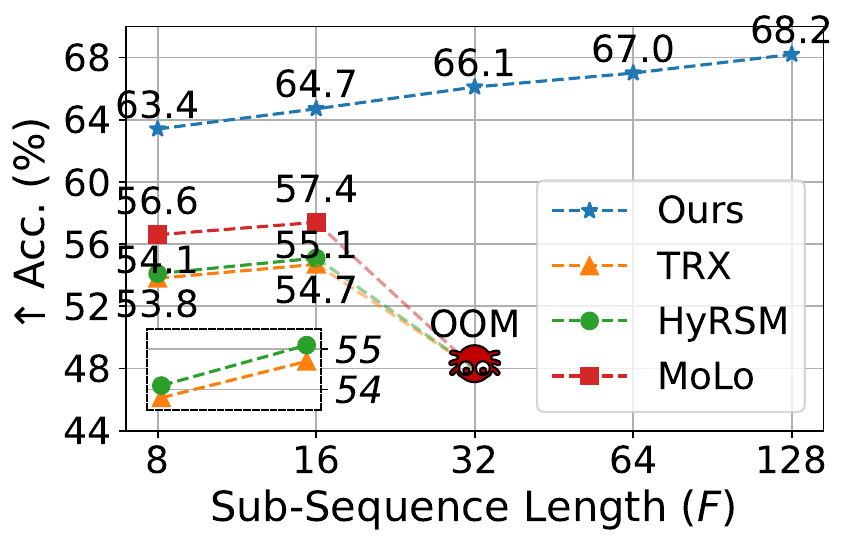}}
	\subfigure[Kinetics]{\includegraphics[width=0.23\textwidth]{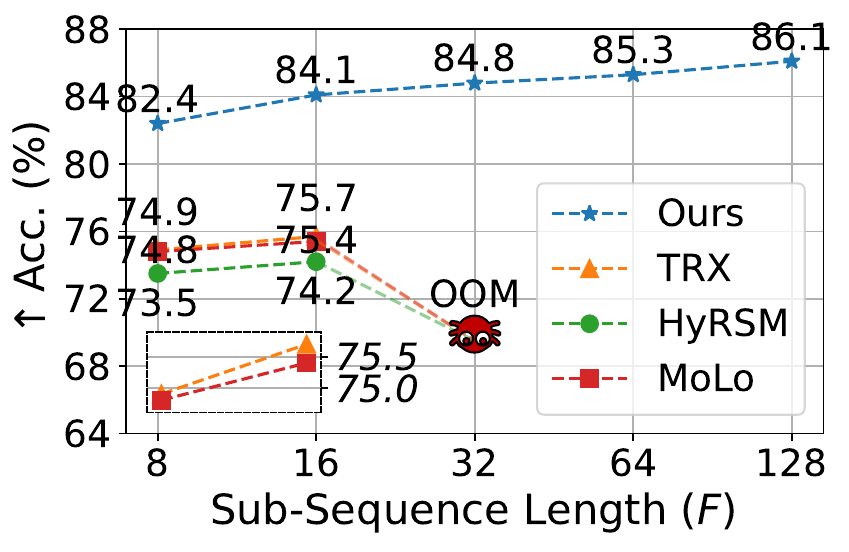}}
	\caption{Curves~($\uparrow$~Acc.~\%, 1-shot setting) on long sub-sequence, where “OOM” means “out of memory”.}
	\label{fig: com long}
\end{figure}

Through increasing $F$, we can get a long sub-sequence evaluation on three recent methods and our Manta. As shown in \figurename~\ref{fig: com long}, a remarkable performance improvement occurs when extending $F$ from 8 to 16. Any further extension of sub-sequence length will cause an “OOM” issue. These phenomena can be observed in all other methods while Manta performs stably even when the length of the sub-sequence is extremely lengthened to 128. We believe the promising performance is from long-sequence modeling and the special design of our Manta.  
\subsubsection{More Complex Tasks.} 
\indent When processing long sub-sequences in FSAR, setting more classes~($N>5$) in each task can enhance the complexity, evaluating the generalization of models. From the results demonstrated in \tablename~\ref{tab: complex}, we can see that performance degrades due to more complex tasks. TRX, HyRSM, and MoLo suffer from a drastic decline in performance while our Manta can maintain a stable level. The above results prove that local feature emphasis, effective temporal alignment, and relieving intra-class variance accumulation are more needed when processing long sub-sequences in more complex tasks.
\begin{table}[t]
	\centering
	\small
	\begin{tabular}{l|l|ccccc}
		\Xhline{1pt}
		\multirow{2}{*}{Datasets} & \multirow{2}{*}{Methods} & \multicolumn{5}{c}{$N$-Way 1-Shot} \\ \cline{3-7} 
		&  & \multicolumn{1}{c|}{6} & \multicolumn{1}{c|}{7} & \multicolumn{1}{c|}{8} & \multicolumn{1}{c|}{9} & 10 \\ \hline
		\multirow{4}{*}{SSv2} & TRX & \multicolumn{1}{c|}{53.8} & \multicolumn{1}{c|}{51.9} & \multicolumn{1}{c|}{49.7} & \multicolumn{1}{c|}{48.3} & 45.7 \\
		& HyRSM & \multicolumn{1}{c|}{53.6} & \multicolumn{1}{c|}{50.4} & \multicolumn{1}{c|}{48.1} & \multicolumn{1}{c|}{45.2} & 42.9 \\
		& MoLo & \multicolumn{1}{c|}{55.2} & \multicolumn{1}{c|}{53.0} & \multicolumn{1}{c|}{50.8} & \multicolumn{1}{c|}{48.5} & 45.1 \\ \cline{2-7} 
		& Manta & \multicolumn{1}{c|}{\textbf{64.1}} & \multicolumn{1}{c|}{\textbf{63.5}} & \multicolumn{1}{c|}{\textbf{62.8}} & \multicolumn{1}{c|}{\textbf{62.2}} & \textbf{61.7} \\
		\hline
		\multirow{4}{*}{Kinetics} & TRX & \multicolumn{1}{c|}{74.5} & \multicolumn{1}{c|}{72.7} & \multicolumn{1}{c|}{70.9} & \multicolumn{1}{c|}{69.4} & 68.2 \\
		& HyRSM & \multicolumn{1}{c|}{73.1} & \multicolumn{1}{c|}{70.2} & \multicolumn{1}{c|}{67.5} & \multicolumn{1}{c|}{64.8} & 62.5 \\
		& MoLo & \multicolumn{1}{c|}{74.3} & \multicolumn{1}{c|}{71.5} & \multicolumn{1}{c|}{72.7} & \multicolumn{1}{c|}{67.1} & 64.2 \\ \cline{2-7} 
		& Manta & \multicolumn{1}{c|}{\textbf{83.9}} & \multicolumn{1}{c|}{\textbf{83.4}} & \multicolumn{1}{c|}{\textbf{82.8}} & \multicolumn{1}{c|}{\textbf{82.2}} & \textbf{81.8} \\
		\Xhline{1pt}
	\end{tabular}
	\caption{Comparison~($\uparrow$~Acc.~\%) under more complex tasks.}
	\label{tab: complex}
\end{table}
\subsubsection{Frame-Level Noise.}
Frame-level noise means a sub-sequence mixed with multiple irrelevant frames~($F^*$). Conducting experiments with noise-level noise can simulate intra-class variance caused by authentic shooting conditions. Such a setting puts forward higher requirements for the robustness of models to face the uncertainty of samples. Results are revealed in \tablename~\ref{tab: f noise}. Intra-class variance accumulated by long sub-sequences has a negative impact. Several performance declines can be observed in TRX, HyRSM, and MoLo while our Manta is significantly less influenced. Therefore, we believe that our special design of Manta can improve the robustness and relieve the negative impact of intra-class variance accumulation.
\begin{table}[ht]
	\centering
	\small
	\begin{tabular}{l|l|ccccc}
		\Xhline{1pt}
		\multirow{2}{*}{Datasets} & \multirow{2}{*}{Methods} & \multicolumn{5}{c}{Noisy Frame Numbers~($F^*$)} \\ \cline{3-7} 
		&  & \multicolumn{1}{c|}{0} & \multicolumn{1}{c|}{2} & \multicolumn{1}{c|}{4} & \multicolumn{1}{c|}{6} & 8 \\ \hline
		\multirow{4}{*}{SSv2} & TRX & \multicolumn{1}{c|}{54.7} & \multicolumn{1}{c|}{51.1} & \multicolumn{1}{c|}{47.9} & \multicolumn{1}{c|}{45.2} & 41.8 \\
		& HyRSM & \multicolumn{1}{c|}{55.1} & \multicolumn{1}{c|}{51.8} & \multicolumn{1}{c|}{48.9} & \multicolumn{1}{c|}{45.7} & 42.1 \\
		& MoLo & \multicolumn{1}{c|}{57.4} & \multicolumn{1}{c|}{54.2} & \multicolumn{1}{c|}{51.4} & \multicolumn{1}{c|}{49.1} & 45.8 \\ \cline{2-7} 
		& Manta & \multicolumn{1}{c|}{\textbf{64.7}} & \multicolumn{1}{c|}{\textbf{63.8}} & \multicolumn{1}{c|}{\textbf{62.6}} & \multicolumn{1}{c|}{\textbf{61.5}} & \textbf{60.1} \\
		\hline
		\multirow{4}{*}{Kinetics} & TRX & \multicolumn{1}{c|}{75.7} & \multicolumn{1}{c|}{72.1} & \multicolumn{1}{c|}{69.0} & \multicolumn{1}{c|}{65.7} & 62.5 \\
		& HyRSM & \multicolumn{1}{c|}{74.2} & \multicolumn{1}{c|}{70.9} & \multicolumn{1}{c|}{67.6} & \multicolumn{1}{c|}{64.7} & 61.5 \\
		& MoLo & \multicolumn{1}{c|}{75.4} & \multicolumn{1}{c|}{72.5} & \multicolumn{1}{c|}{69.4} & \multicolumn{1}{c|}{66.1} & 62.9 \\ \cline{2-7} 
		& Manta & \multicolumn{1}{c|}{\textbf{84.1}} & \multicolumn{1}{c|}{\textbf{82.7}} & \multicolumn{1}{c|}{\textbf{81.5}} & \multicolumn{1}{c|}{\textbf{80.4}} & \textbf{79.1} \\
		\Xhline{1pt}
	\end{tabular}
	\caption{Comparison~($\uparrow$~Acc.~\%) with frame-level noise.}
	\label{tab: f noise}
\end{table} 
\section{Conclusion}
In this paper, we propose Manta, specifically enhancing Mamba for FSAR of long sub-sequence. Manta emphasizes local feature modeling and alignment through Matryoshka Mamba and mitigates the negative impact of intra-class variance accumulation using a hybrid contrastive learning paradigm. Manta achieves new state-of-the-art performance across various benchmarks. Extensive studies demonstrate its competitiveness, generalization, and robustness, particularly for long sub-sequences. We hope our framework will inspire future research in FSAR.

\appendix

\setcounter{figure}{0}
\setcounter{table}{0}
\renewcommand{\thefigure}{\Roman{figure}}
\renewcommand{\thetable}{\Roman{table}}

\section{Supplementary Materials}
In the supplementary materials, all experiments select ResNet-50 as the backbone for long sub-sequence~($F=16$) and we provide:
\begin{itemize}
	\item Study of Mamba Branch.
	\item Study of Contrastive Branch.
	\item Additional Robustness Study.
	\item Computational Complexity. 
\end{itemize}

\appendix

\subsection{Study of Mamba Branch.}
\subsubsection{Base Model.}
\indent Due to the higher accuracy and better efficiency, we select the up-to-date Mamba-2~\cite{dao2024trans} as the base model of our Inner Module and Outer Module. To compare the performance under different base models, we also conducted experiments with Mamba-1~\cite{gu2023mamba}. From the results demonstrated in \tablename~\ref{tab: base model}, we observe that Mamba-2 performs better than Mamba-1. The reason is that Mamba-2 enlarges the state space to 16 or even larger and integrates the attention mechanism~\cite{vaswani2017attention}, resulting in better performance Therefore, we identify Mamba-2 as the base model.
\begin{table}[ht]
	\centering
	\small
	\begin{tabular}{c|c|cc|cc}
		\Xhline{1pt}
		\multirow{2}{*}{IM} & \multirow{2}{*}{OM} & \multicolumn{2}{c|}{SSv2} & \multicolumn{2}{c}{Kinetics} \\ \cline{3-6} 
		&  & \multicolumn{1}{c|}{1-shot} & 5-shot & \multicolumn{1}{c|}{1-shot} & 5-shot \\ \hline
		M-1 & M-1 & \multicolumn{1}{c|}{64.1} & 87.6 & \multicolumn{1}{c|}{83.7} & 95.1 \\
		M-1 & M-2 & \multicolumn{1}{c|}{63.9} & 87.8 & \multicolumn{1}{c|}{83.8} & 95.4 \\
		M-2 & M-1 & \multicolumn{1}{c|}{64.2} & 88.1 & \multicolumn{1}{c|}{83.6} & 95.8 \\
		M-2 & M-2 & \multicolumn{1}{c|}{\textbf{64.7}} & \textbf{88.7} & \multicolumn{1}{c|}{\textbf{84.1}} & \textbf{96.2} \\ 
		\Xhline{1pt}
	\end{tabular}
	\caption{Comparison~($\uparrow$ Acc.~\%) between Mamba-1~(M-1) and Mamba-2~(M-2).  Few-shot setting: 5-way 1-shot.}
	\label{tab: base model}
\end{table}  
\subsubsection{Shared Parameters.}
\indent Inner Module is applied for reinforcing local features while Outer Module is designed for executing implicit temporal alignment. Exploring independent and shared parameters between bub-branches can improve the differentiation of their functions. Therefore, we complete experiments by various forms of the parameters and list the results in \tablename~\ref{tab: share}. Using all independent or shared parameters does not provide the best performance. Inner Module with independent parameters and Outer Module with shared parameters can achieve the best performance. We suppose that the variable local features need more flexible bidirectional scanning from Inner Module with independent parameters. While temporal alignment is a bidirectional symmetric structure, Outer Module with shared parameters can make alignment learned in both directions reinforce each other.
\begin{table}[t]
	\centering
	\small
	\begin{tabular}{c|c|cc|cc}
		\Xhline{1pt}
		\multirow{2}{*}{IM} & \multirow{2}{*}{OM} & \multicolumn{2}{c|}{SSv2} & \multicolumn{2}{c}{Kinetics} \\ \cline{3-6} 
		&  & \multicolumn{1}{c|}{1-shot} & 5-shot & \multicolumn{1}{c|}{1-shot} & 5-shot \\ \hline
		I-P & I-P & \multicolumn{1}{c|}{63.2} & 87.1 & \multicolumn{1}{c|}{83.5} & 95.1 \\
		I-P & S-P & \multicolumn{1}{c|}{\textbf{64.7}} & \textbf{88.7} & \multicolumn{1}{c|}{\textbf{84.1}} & \textbf{96.2} \\
		S-P & I-P & \multicolumn{1}{c|}{64.1} & 86.9 & \multicolumn{1}{c|}{83.4} & 95.5 \\
		S-P & S-P & \multicolumn{1}{c|}{63.9} & 87.3 & \multicolumn{1}{c|}{83.2} & 95.8 \\ \Xhline{1pt}
	\end{tabular}
	\caption{Comparison~($\uparrow$ Acc.~\%) between independent~(I-P) and shared parameters~(S-P).  Few-shot setting: 5-way 1-shot.}
	\label{tab: share}
\end{table} 
\subsubsection{Learnable Weights $w_{o}$.}
\indent To discuss the difference between various methods of multi-scale combination, we compare fixed and learnable weights in our experiments. From the results revealed in \tablename~\ref{tab: learnable}, we find that applying learnable weights~($w_{o}$~\ding{51}) performs better than directly using fixed ones~($w_{o}$~\ding{55}). The reason is that weights allow the model to dynamically adjust appropriate significance to the features at each scale, thus enhancing the overall feature representation. 
\begin{table}[ht]
	\centering
	\small
	\begin{tabular}{c|cc|cc}
		\Xhline{1pt}
		\multirow{2}{*}{$w_{o}$} & \multicolumn{2}{c|}{SSv2} & \multicolumn{2}{c}{Kinetics} \\ \cline{2-5} 
		& \multicolumn{1}{c|}{1-shot} & 5-shot & \multicolumn{1}{c|}{1-shot} & 5-shot \\ \hline
		\ding{51} & \multicolumn{1}{c|}{\textbf{64.7}} & \textbf{88.7} & \multicolumn{1}{c|}{\textbf{84.1}} & \textbf{96.2} \\
		\ding{55} & \multicolumn{1}{c|}{62.2} & 86.1 & \multicolumn{1}{c|}{81.3} & 92.9 \\ \Xhline{1pt}
	\end{tabular}
	\caption{Comparison~($\uparrow$~Acc.~\%) between fixed and learnable weights $w_{o}$. Few-shot setting: 5-way 1-shot.}
	\label{tab: learnable}
\end{table} 
\subsubsection{Temporal Order.}
\indent We intuitively believe that temporal order is crucial for feature representation in FSAR. To verify this conjecture, we reverse the order of the support set while maintaining the temporal order of the query, and then conduct experiments for comparison. The results demonstrated in \tablename~\ref{tab: temporal} show that all datasets have a degeneration with reserved order. It is worth noting that the drop in SSv2 is more significant than Kinetics. The dataset type contributes to our observation that temporal-related SSv2 is more sensitive to changes in temporal order. Therefore, the negative impact of reserved order is smaller on Kinetics.
\begin{table}[ht]
	\centering
	\small
	\begin{tabular}{c|cc|cc}
		\Xhline{1pt}
		\multirow{2}{*}{\begin{tabular}[c]{@{}l@{}}Reversed\\ Order\end{tabular}} & \multicolumn{2}{c|}{SSv2} & \multicolumn{2}{c}{Kinetics} \\ \cline{2-5} 
		& \multicolumn{1}{c|}{1-shot} & 5-shot & \multicolumn{1}{c|}{1-shot} & 5-shot \\ \hline
		\ding{51} & \multicolumn{1}{c|}{58.2} & 76.3 & \multicolumn{1}{c|}{78.2} & 90.1 \\
		\ding{55} & \multicolumn{1}{c|}{\textbf{64.7}} & \textbf{88.7} & \multicolumn{1}{c|}{\textbf{84.1}} & \textbf{96.2} \\ \Xhline{1pt}
	\end{tabular}
	\caption{Comparison~($\uparrow$~Acc.~\%) between reserved and temporal order. Few-shot setting: 5-way 1-shot.}
	\label{tab: temporal}
\end{table} 
\subsubsection{Weight Factor $\lambda$ of Loss.}
\indent The weight factor $\lambda$ of loss is designed to adjust values of $\mathcal{L}_{\text{ce}}$ to the same level of $\mathcal{L}_{\text{hc}}$. We aim to find the most suitable $\lambda$ from extensive experiments. The results are recorded in \tablename~\ref{tab: factor}. With the increase of $\lambda$, the performance gradually improves. However, we observe that the best results always be achieved when $\lambda=4$. If the value of $\lambda$ continues to grow, there will be a small decline in performance. Therefore, we determined that $\lambda=4$ is the most appropriate weight factor.
\begin{table}[ht]
	\centering
	\small
	\begin{tabular}{c|cc|cc}
		\Xhline{1pt}
		\multirow{2}{*}{$\lambda$} & \multicolumn{2}{c|}{SSv2} & \multicolumn{2}{c}{Kinetics} \\ \cline{2-5} 
		& \multicolumn{1}{c|}{1-shot} & 5-shot & \multicolumn{1}{c|}{1-shot} & 5-shot \\ \hline
		$\lambda=1$ & \multicolumn{1}{c|}{60.8} & 86.4 & \multicolumn{1}{c|}{81.8} & 92.0 \\
		$\lambda=2$ & \multicolumn{1}{c|}{62.6} & 87.3 & \multicolumn{1}{c|}{82.4} & 93.7 \\
		$\lambda=3$ & \multicolumn{1}{c|}{63.9} & 88.0 & \multicolumn{1}{c|}{82.9} & 94.5 \\ 
		$\lambda=4$ & \multicolumn{1}{c|}{\textbf{64.7}} & \textbf{88.7} & \multicolumn{1}{c|}{\textbf{84.1}} & \textbf{96.2} \\ 
		$\lambda=5$ & \multicolumn{1}{c|}{64.1} & 88.1 & \multicolumn{1}{c|}{82.9} & 94.6 \\ 
		$\lambda=6$ & \multicolumn{1}{c|}{63.7} & 87.8 & \multicolumn{1}{c|}{82.3} & 94.2 \\  
		\Xhline{1pt}
	\end{tabular}
	\caption{Comparison~($\uparrow$ Acc.~\%) of various weight factor $\lambda$. }
	\label{tab: factor}
\end{table} 
\subsection{Study of Contrastive Branch.}
\subsubsection{Unsupervised $z^\text{p}$ and $z^\text{n}$ Construction.}
\indent In unsupervised contrastive learning of FSAR, we hypothesis that support set $\mathcal{S}$ and query set $\mathcal{Q}$ all have $N\times K$ samples for a simple presentation.
\begin{equation*}
	\begin{aligned}
		\mathcal{S}&=\left\{ \left( sx_{1}^{1},\dots ,sx_{K}^{1} \right) ,\dots ,\left( sx_{1}^{N},\dots ,sx_{K}^{N} \right) \right\}, \\
		\mathcal{Q}&=\left\{ \left( qx_{1}^{1},\dots ,qx_{K}^{1} \right) ,\dots ,\left( qx_{1}^{N},\dots ,qx_{K}^{N} \right) \right\}.
	\end{aligned}
\end{equation*}
For contrastive loss $\mathcal{L}_{\text{con}}^{\mathcal{Q}}$, we randomly select a sample $qx_{k}^{n} \left[ \left( n = 1, \cdots, N \right) \land \left(k = 1, \cdots, K \right) \right]$ as an anchor vector. Other $K-1$ samples of the same class are $z^\text{p}$ while the remaining $\left(N-1 \right)\times K$ samples are $z^\text{n}$. When it comes to unsupervised $\mathcal{L}_{\text{con}}^{\mathcal{SQ}}$, we form $\mathcal{S}$ and $\mathcal{Q}$ into a larger set $\mathcal{SQ}=\mathcal{S}\cup \mathcal{Q}$ with $2N\times K$ samples: 
\begin{equation*}
	\begin{aligned}
		\mathcal{SQ}=\left\{ \left( x_{1}^{1},\dots ,x_{2K}^{1} \right) ,\dots ,\left( x_{1}^{\left( N \right)},\dots ,x_{2K}^{\left( N \right)} \right) \right\} .
	\end{aligned}
\end{equation*}
A random sample $x_{k}^{n} \left[ \left( n = 1, \cdots, N \right) \land \left(k = 1, \cdots, K \right) \right]$ is selected as an anchor vector in the same way. Therefore, $2K-1$ samples of the same class are denoted as $z^\text{p}$, and the remaining $\left(N-1\right) \times 2K$ samples are denoted as $z^\text{n}$. Following the paradigm of contrastive learning loss, $\mathcal{L}_{\text{con}}^{\mathcal{Q}}$ and $\mathcal{L}_{\text{con}}^{\mathcal{SQ}}$ can be calculated.
\subsubsection{$\mathcal{L}_{\text{con}}^{\mathcal{S}}$, $\mathcal{L}_{\text{con}}^{\mathcal{Q}}$ and $\mathcal{L}_{\text{con}}^{\mathcal{SQ}}$.}
\indent After obtaining $\mathcal{L}_{\text{hc}}$ from $\mathcal{L}_{\text{con}}^{\mathcal{S}}$, $\mathcal{L}_{\text{con}}^{\mathcal{Q}}$, and $\mathcal{L}_{\text{con}}^{\mathcal{SQ}}$, we aim to explore the effect of each component. By analyzing $\mathcal{L}_{\text{hc}}$ in detail, we conducted experiments to evaluate the impact of each contrastive loss term. Results presented in \tablename~\ref{tab: contrastive} reveal that using any single type of contrastive learning is beneficial for performance improvement. However, employing both contrastive learning losses together yields better results than using just one. Finally, applying the full $\mathcal{L}_{\text{hc}}$ in a hybrid manner achieves the best performance. This observation confirms that all components of $\mathcal{L}_{\text{hc}}$ are necessary.
\begin{table}[ht]
	\centering
	\small
	\begin{tabular}{c|c|c|cc|cc}
		\Xhline{1pt}
		\multirow{2}{*}{$\mathcal{L}_{\text{con}}^{\mathcal{S}}$} & \multirow{2}{*}{$\mathcal{L}_{\text{con}}^{\mathcal{Q}}$} & \multirow{2}{*}{$\mathcal{L}_{\text{con}}^{\mathcal{SQ}}$} & \multicolumn{2}{c|}{SSv2} & \multicolumn{2}{c}{Kinetics} \\ \cline{4-7} 
		&  &  & \multicolumn{1}{c|}{1-shot} & 5-shot & \multicolumn{1}{c|}{1-shot} & 5-shot \\ \hline
		\ding{51} & \ding{55} & \ding{55} & \multicolumn{1}{c|}{63.3} & 87.1 & \multicolumn{1}{c|}{81.6} & 93.2 \\
		\ding{55} & \ding{51} & \ding{55} & \multicolumn{1}{c|}{62.6} & 86.8 & \multicolumn{1}{c|}{81.1} & 92.7 \\
		\ding{55} & \ding{55} & \ding{51} & \multicolumn{1}{c|}{64.2} & 87.4 & \multicolumn{1}{c|}{81.9} & 93.7 \\ \hline
		\ding{51} & \ding{51} & \ding{55} & \multicolumn{1}{c|}{64.9} & 87.7 & \multicolumn{1}{c|}{82.3} & 94.1 \\
		\ding{51} & \ding{55} & \ding{51} & \multicolumn{1}{c|}{64.1} & 88.0 & \multicolumn{1}{c|}{82.9} & 94.8 \\
		\ding{55} & \ding{51} & \ding{51} & \multicolumn{1}{c|}{63.5} & 87.6 & \multicolumn{1}{c|}{82.5} & 94.3 \\ \hline
		\ding{51} & \ding{51} & \ding{51} & \multicolumn{1}{c|}{\textbf{64.7}} & \textbf{88.7} & \multicolumn{1}{c|}{\textbf{84.1}} & \textbf{96.2} \\ 
		\Xhline{1pt}
	\end{tabular}
	\caption{Comparison~($\uparrow$ Acc.~\%) of various contrastive loss. }
	\label{tab: contrastive}
\end{table} 
\subsubsection{Temperature $\tau$.}
\indent After determining the hybrid way of $\mathcal{L}_{\text{hc}}$, the impact of hyper-parameter temperature $\tau$ is also discussed. By adjusting $\tau$ from 0.01 to 0.11, we conducted several experiments and reported results in \tablename~\ref{tab: temperature}. The overall trend of performance is gradually improved with the $\tau$ growth. However, the improvement is not unbounded, peaking at $\tau=0.07$ and saturating. Further growth of $\tau$ makes the performance drop. Therefore, the temperature $\tau$ is determined to be 0.07.
\begin{table}[t]
	\centering
	\small
	\begin{tabular}{c|cc|cc}
		\Xhline{1pt}
		\multirow{2}{*}{$\tau$} & \multicolumn{2}{c|}{SSv2} & \multicolumn{2}{c}{Kinetics} \\ \cline{2-5} 
		& \multicolumn{1}{c|}{1-shot} & 5-shot & \multicolumn{1}{c|}{1-shot} & 5-shot \\ \hline
		$\tau=0.01$ & \multicolumn{1}{c|}{63.5} & 87.6& \multicolumn{1}{c|}{82.4} & 94.4 \\
		$\tau=0.03$ & \multicolumn{1}{c|}{63.8} & 87.9 & \multicolumn{1}{c|}{82.7} & 94.6 \\
		$\tau=0.05$ & \multicolumn{1}{c|}{64.1} & 88.2 & \multicolumn{1}{c|}{83.1} & 94.9 \\ 
		$\tau=0.07$ & \multicolumn{1}{c|}{\textbf{64.7}} & \textbf{88.7} & \multicolumn{1}{c|}{\textbf{84.1}} & \textbf{96.2} \\ 
		$\tau=0.09$ & \multicolumn{1}{c|}{64.2} & 88.3 & \multicolumn{1}{c|}{83.2} & 95.0 \\ 
		$\tau=0.11$ & \multicolumn{1}{c|}{63.9} & 88.0 & \multicolumn{1}{c|}{82.9} & 94.8 \\  
		\Xhline{1pt}
	\end{tabular}
	\caption{Comparison~($\uparrow$ Acc.~\%) of various temperature $\tau$. }
	\label{tab: temperature}
\end{table}
\subsection{Additional Robustness Study.}
\subsubsection{The t-SNE Visualization.}
\begin{figure}[ht]
	\centering
	\subfigure[w/o Manta]{\includegraphics[width=0.18\textwidth]{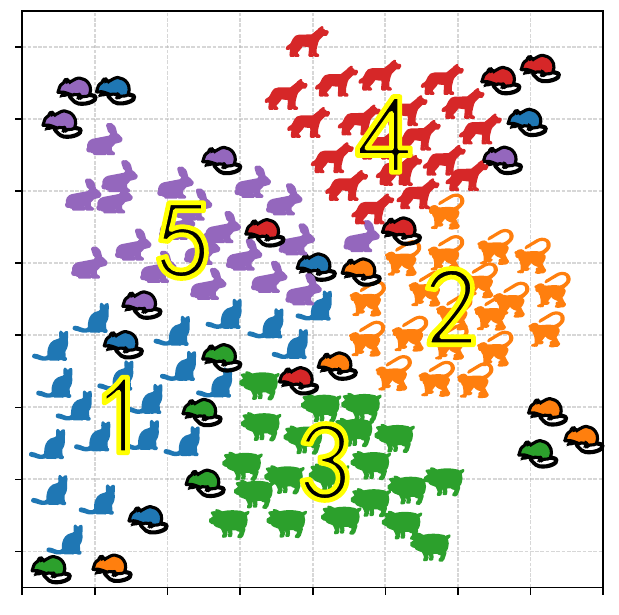}}
	\subfigure[w/ Manta]{\includegraphics[width=0.18\textwidth]{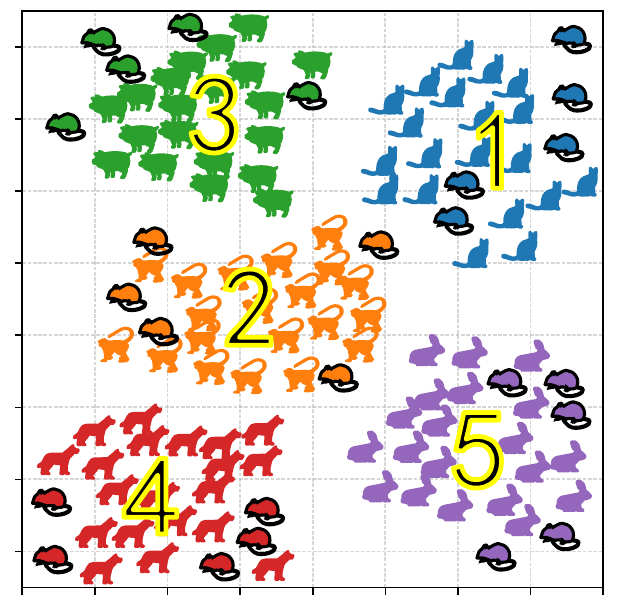}}
	\caption{Challenging action classes are selected from Kinetics. Blue: “Unboxing”, Orange: “Folding paper”, Green: “Hurling (sport)”, Red: “Tap dancing”, Purple: “Throwing axe”. In the t-SNE visualization, points with the same color are samples of the same class, and points with black borders are samples with large intra-class variance. Few-shot setting: 5-way 25-shot.}
	\label{fig: tsen}
\end{figure}

\indent From the testing set of Kinetics, 5 challenging action classes with similar appearances including “Unboxing”, “Folding paper”, “Hurling (sport)”, “Tap dancing”, and “Throwing axe” is selected. In each class, there are several samples with significant intra-class variance. From the t-SNE~\cite{van2008visualizing}, feature representation without Manta~(\figurename~\ref{fig: tsen}(a)) has overlap between classes because of the similar visual features of actions. Due to the negative impact of intra-class variance accumulation, many samples are separated and cannot be clustered well. The above phenomena can be alleviated with the help of Manta~(\figurename~\ref{fig: tsen}(b)). Specifically, the different classes are separated and the samples belonging to the same class are tightly clustered. For these independently separated samples with significant intra-class variances, a tendency of clustering to their corresponding class is highly remarkable.    
\subsubsection{Confusion Matrix.}
\begin{figure}[t]
	\centering
	\subfigure[w/o Manta~($\uparrow$ Acc. 55.6\%)]{\includegraphics[width=0.16\textwidth]{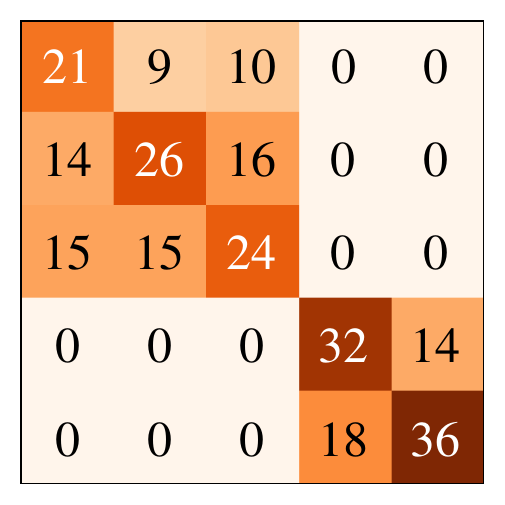}}
	\subfigure[w/ Manta~($\uparrow$ Acc. 89.2\%)]{\includegraphics[width=0.16\textwidth]{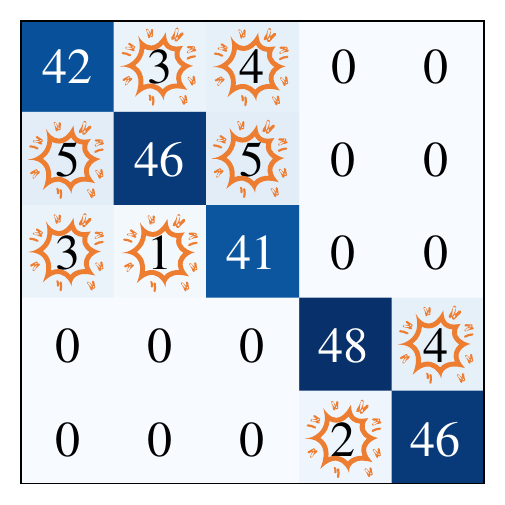}}
	\caption{Confusion Matrices of 250 samples from 5 action classes. Ground truth $y_{Q}^{j}$: $X$-axis, predict labels $\tilde{y}_{Q}^{j}$: $Y$-axis. From top to bottom and left to right, action class are “High jump”, “Basketball”, “Basketball dunk”, “Soccer juggling”, and “Soccer penalty” from UCF101. Misclassification be reduced to a low level are highlighted.}
	\label{fig: matrix}
\end{figure}

\indent Although long sub-sequence has inherent advantage in expressing entire actions, directly applying Mamba for FSAR of long sub-sequence cannot obtain satisfactory performance We select 250 samples from 5 classes of UCF101 and visualize the effect of Manta by confusion matrix. As illustrated in \figurename~\ref{fig: matrix}, “Basketball”  and “Basketball dunk” can be seen as actions cover “High jump”. “Soccer juggling”, and “Soccer penalty” can be treated as similar composite actions. We observe Manta reduces the misclassification between them to a lower level. The accuracy of these classes are improved from 55.6\% to 89.2\%. The experimental results demonstrates the superior of Manta for FSAR of long sub-sequence.
\subsubsection{Sample-Level Noise.}
\indent During the collection of a large number of samples in the real environment, sample-level noise inevitably occurs for different reasons. Sometimes samples from other classes might be mixed into a particular class. It is time-consuming and laborious to correct them. Therefore evaluating with frame-level noise can simulate more authentic scenarios, testing the robustness of models. As shown in \tablename~\ref{tab: s noise}, we find that the performance of TRX~\cite{perrett2021temporal}, HyRSM~\cite{wang2022hybrid}, and MoLo~\cite{wang2023molo} decrease severely with the sample-level noise ratio increases. Our Manta is less negatively affected and performs smoothly under sample-level noise. This experiment proves that emphasizing local features, executing temporal alignment, and relieving intra-class variance accumulation can improve robustness.
\begin{table}[ht]
	\centering
	\small
	\begin{tabular}{l|l|ccccc}
		\Xhline{1pt}
		\multirow{2}{*}{Datasets} & \multirow{2}{*}{Methods} & \multicolumn{5}{c}{Sample-Level Noise Ratio} \\ \cline{3-7} 
		&  & \multicolumn{1}{c|}{0\%} & \multicolumn{1}{c|}{10\%} & \multicolumn{1}{c|}{20\%} & \multicolumn{1}{c|}{30\%} & 40\% \\ \hline
		\multirow{4}{*}{SSv2} & TRX & \multicolumn{1}{c|}{74.6} & \multicolumn{1}{c|}{71.8} & \multicolumn{1}{c|}{68.6} & \multicolumn{1}{c|}{65.8} & 62.7 \\
		& HyRSM & \multicolumn{1}{c|}{75.1} & \multicolumn{1}{c|}{72.6} & \multicolumn{1}{c|}{69.4} & \multicolumn{1}{c|}{66.1} & 62.9 \\
		& MoLo & \multicolumn{1}{c|}{75.8} & \multicolumn{1}{c|}{73.4} & \multicolumn{1}{c|}{71.2} & \multicolumn{1}{c|}{68.9} & 65.8 \\ \cline{2-7} 
		& Manta & \multicolumn{1}{c|}{\textbf{91.6}} & \multicolumn{1}{c|}{\textbf{90.4}} & \multicolumn{1}{c|}{\textbf{89.2}} & \multicolumn{1}{c|}{\textbf{87.8}} & \textbf{86.7} \\
		\hline
		\multirow{4}{*}{Kinetics} & TRX & \multicolumn{1}{c|}{89.8} & \multicolumn{1}{c|}{85.9} & \multicolumn{1}{c|}{83.1} & \multicolumn{1}{c|}{80.3} & 77.4 \\
		& HyRSM & \multicolumn{1}{c|}{90.7} & \multicolumn{1}{c|}{88.2} & \multicolumn{1}{c|}{85.8} & \multicolumn{1}{c|}{83.6} & 80.5 \\
		& MoLo & \multicolumn{1}{c|}{91.2} & \multicolumn{1}{c|}{88.4} & \multicolumn{1}{c|}{85.6} & \multicolumn{1}{c|}{82.2} & 80.6 \\ \cline{2-7} 
		& Manta & \multicolumn{1}{c|}{\textbf{98.5}} & \multicolumn{1}{c|}{\textbf{97.2}} & \multicolumn{1}{c|}{\textbf{95.7}} & \multicolumn{1}{c|}{\textbf{94.5}} & \textbf{93.3} \\
		\Xhline{1pt}
	\end{tabular}
	\caption{Comparison~($\uparrow$~Acc.~\%) with sample-level noise.  Few-shot setting: 5-way 10-shot.}
	\label{tab: s noise}
\end{table} 
\subsubsection{Background Noise.}
\begin{figure}[ht]
	\centering
	\includegraphics[width=0.46\textwidth]{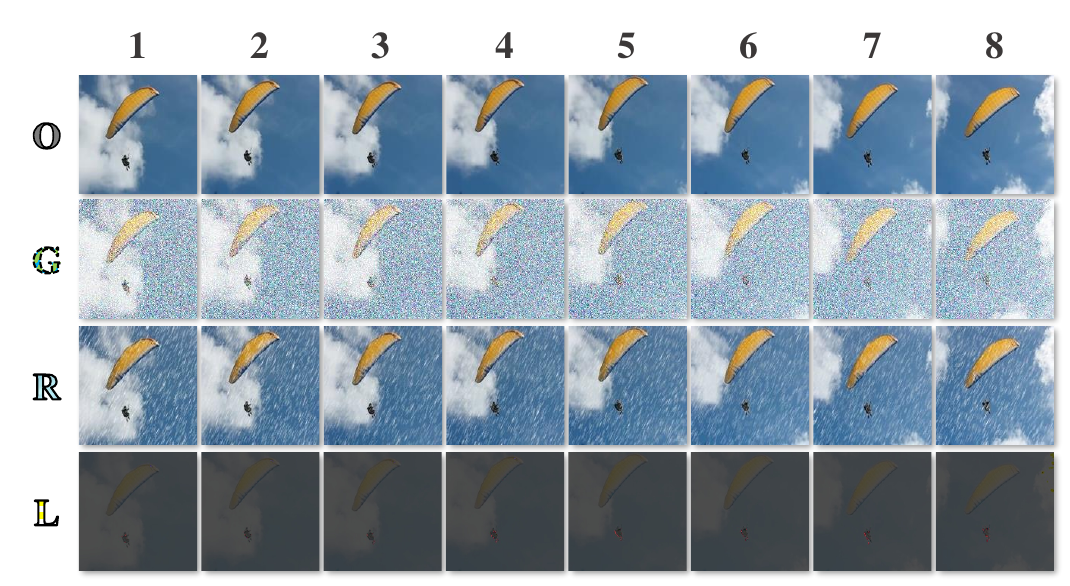} 
	\caption{Example “Paragliding” from Kinetics for background noise demonstration. Original frames~(O) are located in the first row. Gaussian noise~(G), rainy noise~(R), and light noise~(L) are added in the second, third, and fourth rows.}
	\label{fig: background}
\end{figure}

\indent The real shooting condition not only imposes frame-level noise into samples but also introduces different types of background noise. To be specific, Gaussian noise is related to issues of hardware. Rainy noise is caused by changeable weather. Changes in lighting conditions bring light noise. These all challenge the robustness of the model. Therefore, we add Gaussian, rainy, and light noise to 25\% samples respectively to create more intra-class variances, as depicted in \figurename~\ref{fig: background}. The results are listed in \tablename~\ref{tab: b noise}. Consistent with our intuition, Gaussian, rainy, and light noise all negatively affect the performance of the model. Taking the most significant influence from light noise as an example, we observe that our Manta experiences an about 5\% performance decline. Compared with other recent methods, Manta performs much more robustly. A similar phenomenon can also be observed in other types of background noise. 
\begin{table}[ht]
	\centering
	\small
	\begin{tabular}{l|l|cccc}
		\Xhline{1pt}
		\multirow{2}{*}{Datasets} & \multirow{2}{*}{Methods} & \multicolumn{4}{c}{Background Noise Type} \\ \cline{3-6} 
		&  & \multicolumn{1}{c|}{O} & \multicolumn{1}{c|}{G} & \multicolumn{1}{c|}{R} & L \\ \hline
		\multirow{4}{*}{SSv2} & TRX & \multicolumn{1}{c|}{54.7} & \multicolumn{1}{c|}{52.8} & \multicolumn{1}{c|}{51.5} & 42.0 \\
		& HyRSM & \multicolumn{1}{c|}{55.1} & \multicolumn{1}{c|}{53.7} & \multicolumn{1}{c|}{52.5} & 43.5 \\
		& MoLo & \multicolumn{1}{c|}{57.4} & \multicolumn{1}{c|}{55.9} & \multicolumn{1}{c|}{54.7} & 44.2 \\ \cline{2-6} 
		& Manta & \multicolumn{1}{c|}{\textbf{64.7}} & \multicolumn{1}{c|}{\textbf{64.2}} & \multicolumn{1}{c|}{\textbf{63.9}} & \textbf{59.8} \\ \hline
		\multirow{4}{*}{Kinetics} & TRX & \multicolumn{1}{c|}{75.7} & \multicolumn{1}{c|}{74.0} & \multicolumn{1}{c|}{72.6} & 63.3 \\
		& HyRSM & \multicolumn{1}{c|}{74.2} & \multicolumn{1}{c|}{72.3} & \multicolumn{1}{c|}{70.8} & 62.8 \\
		& MoLo & \multicolumn{1}{c|}{75.4} & \multicolumn{1}{c|}{73.3} & \multicolumn{1}{c|}{71.9} & 64.3 \\ \cline{2-6} 
		& Manta & \multicolumn{1}{c|}{\textbf{84.1}} & \multicolumn{1}{c|}{\textbf{83.4}} & \multicolumn{1}{c|}{\textbf{83.0}} & \textbf{78.9} \\ 
		\Xhline{1pt}
	\end{tabular}
	\caption{Comparison~($\uparrow$~Acc.~\%) with original frames~(O), Gaussian noise~(G), rainy noise~(R), and light noise~(L). Few-shot setting: 5-way 1-shot. }
	\label{tab: b noise}
\end{table} 
\subsubsection{Cross Dataset Testing.}
\indent Due to different data distributions, training and testing on various datasets can simulate the performance of models in practical applications. Therefore, we apply Kinetics and SSv2 for cross-dataset testing after eliminating overlapping classes in the training, validation, and test sets. From the results demonstrated in \tablename~\ref{tab: cdt}, our Manta with cross-dataset testing is still ahead of other methods, although this particular test setup can degrade performance. This trend is consistent with regular testing. Compared with other methods, we further find that the performance decline caused by cross-dataset testing is not evident on Manta, proving the better robustness with different data distributions. 
\begin{table}[t]
	\centering
	\small
	\begin{tabular}{l|cc|cc}
		\Xhline{1pt}
		\multirow{2}{*}{Methods} & \multicolumn{2}{c|}{KI$\rightarrow$SS \scriptsize(SS$\rightarrow$SS)} & \multicolumn{2}{c}{SS$\rightarrow$KI \scriptsize(KI$\rightarrow$KI)} \\ \cline{2-5} 
		& \multicolumn{1}{c|}{1-shot} & 5-shot & \multicolumn{1}{c|}{1-shot} & 5-shot \\ \hline
		TRX & \multicolumn{1}{c|}{52.3 \scriptsize(54.7)} & 67.5 \scriptsize(70.8) & \multicolumn{1}{c|}{71.7 \scriptsize(75.7)} & 80.2 \scriptsize(85.9) \\
		HyRSM & \multicolumn{1}{c|}{51.2 \scriptsize(55.1)} & 66.9 \scriptsize(70.7) & \multicolumn{1}{c|}{70.8 \scriptsize(74.2)} & 80.9 \scriptsize(86.2) \\
		MoLo & \multicolumn{1}{c|}{54.3 \scriptsize(57.4)} & 69.4 \scriptsize(72.7) & \multicolumn{1}{c|}{72.1 \scriptsize(75.4)} & 81.3 \scriptsize(85.7) \\ \hline
		Manta & \multicolumn{1}{c|}{\textbf{63.2} \scriptsize\textbf{(64.7)}} & \textbf{88.3} \scriptsize\textbf{(89.4)} & \multicolumn{1}{c|}{\textbf{82.4} \scriptsize\textbf{(84.1)}} & \textbf{92.8 \scriptsize\textbf{(94.2)}} \\ 
		\Xhline{1pt}
	\end{tabular}
	\caption{Comparison~($\uparrow$~Acc.~\%) with cross dataset~(large fonts) and regular testing~(small fonts in brackets). KI$\rightarrow$SS: Kinetics training while SSv2 testing, SS$\rightarrow$KI: SSv2 training while Kinetics testing, SS$\rightarrow$SS: training and testing both on SSv2, KI$\rightarrow$KI: training and testing both on Kinetics.}
	\label{tab: cdt}
\end{table} 
\subsection{Computational Complexity.}
\subsubsection{Inference Speed on Single GPU.}
\indent Inference with a single GPU can help evaluate the model in practical application with limited resources. To realize such hardware conditions, we select only one 24GB NVIDIA GeForce RTX 3090 GPU in a server for 10,000 tasks in this experiment. As results demonstrated in \tablename~\ref{tab: single}, the inference speed of Manta is much faster than other methods based on Transformer. The lower latency of Manta in practice is due to Mamba not only having a more lightweight architecture but also using parallel algorithms optimized for modern hardware, especially GPUs.
\begin{table}[ht]
	\centering
	\small
	\begin{tabular}{l|cc|cc}
		\Xhline{1pt}
		\multirow{2}{*}{Methods} & \multicolumn{2}{c|}{SSv2} & \multicolumn{2}{c}{Kinetics} \\ \cline{2-5} 
		& \multicolumn{1}{c|}{1-shot} & 5-shot & \multicolumn{1}{c|}{1-shot} & 5-shot \\ \hline
		TRX & \multicolumn{1}{c|}{8.67} & 8.83 & \multicolumn{1}{c|}{8.74} & 8.96 \\
		HyRSM & \multicolumn{1}{c|}{6.33} & 6.46 & \multicolumn{1}{c|}{6.86} & 6.92 \\
		MoLo & \multicolumn{1}{c|}{7.83} & 8.02 & \multicolumn{1}{c|}{7.64} & 8.14 \\ \hline
		Manta & \multicolumn{1}{c|}{\textbf{4.25}} & \textbf{4.61} & \multicolumn{1}{c|}{\textbf{4.42}} & \textbf{4.56} \\ 
		\Xhline{1pt}
	\end{tabular}
	\caption{Inference speed~($\downarrow$~hour) with 10,000 random tasks on single 24GB NVIDIA GeForce RTX 3090 GPU.}
	\label{tab: single}
\end{table} 
\subsubsection{Model Structure.}
\indent For a better comprehension of our Manta, we list the major tensor changes from the Mamba branch in \tablename~\ref{tab: structure}. The notations are consistent with the main paper. These can demonstrate how Manta works, helping select appropriate hyper-parameters and model settings. According to abundant experiments conducted in previous sections, we determine hyper-parameters such as multi-scale~($\mathcal{O}=\left\{ 1,2,4 \right\}$), temperature~($\tau=0.07$) and weight factor of loss~($\lambda=4$). Model settings including $\mathrm{IM}\left( \cdot \right)$ with independent parameters, $\mathrm{OM}\left( \cdot \right)$ with shared parameters, applying learnable weights $w_o$, and hybrid contrastive learning loss $\mathcal{L}_{\text{hc}}$ are more clear. The following pseudo-code provides a better analysis of computational complexity.
\begin{table}[ht]
	\centering
	\small
    \setlength{\tabcolsep}{1.8pt}
	\begin{tabular}{l|c|r|c|r}
		\Xhline{1pt}
		Operation & Input & \multicolumn{1}{c|}{Input Size} & Output & \multicolumn{1}{c}{Onput Size} \\ \hline
		\multirow{2}{*}{$f_{\theta}\left( \cdot \right)$} & $S^{ck}$ & [$F$,$C$,$H$,$W$] & ${S}_{f}^{ck}$ & [$F$,$D$] \\
		& $Q^{r}$ & [$F$,$C$,$H$,$W$] & $Q_{f}^{r}$ & [$F$,$D$] \\ \hline
		\multirow{2}{*}{$\mathrm{IM}\left( \cdot \right)$} & ${s}_{f_o}^{ck}$ & [$D$,$o$] & $\tilde{S}_{f_o}^{ck}$ & [$F$,$D$] \\
		& $q_{f_o}^{r}$ & [$D$,$o$] & $\tilde{Q}_{f_o}^{r}$ & [$F$,$D$] \\ \hline
		\multirow{2}{*}{$\mathrm{OM}\left( \cdot \right)$} & $\tilde{S}_{f_o}^{ck}$ & [$F$,$D$] & $\mathrm{OM}\left( \tilde{S}_{f_o}^{ck} \right)$ & [$F$,$D$] \\
		& $\tilde{Q}_{f_o}^{r}$ & [$F$,$D$] & $\mathrm{OM}\left( \tilde{Q}_{f_o}^{r} \right)$ & [$F$,$D$] \\ \hline
		\multirow{2}{*}{Eqn.~(5)} & $\mathring{S}_{f_o}^{ck}$ & [$F$,$D$] & $w_{o}^{S}$ & [$F$,$D$] \\
		& $\mathring{Q}_{f_o}^{r}$ & [$F$,$D$] & $w_{o}^{Q}$ & [$F$,$D$] \\ \hline
		\multirow{4}{*}{Eqn.~(6)} & $w_{o}^{S}$ & [$F$,$D$] & \multirow{2}{*}{$\mathring{S}_{f_o}^{ck}$} & \multirow{2}{*}{[$F$,$D$]} \\
		& $\mathrm{OM}\left( \tilde{S}_{f_o}^{ck} \right)$ & [$F$,$D$] &  &  \\ \cline{2-5}
		& $w_{o}^{Q}$ & [$F$,$D$] & \multirow{2}{*}{$\mathring{Q}_{f_o}^{r}$} & \multirow{2}{*}{[$F$,$D$]} \\
		& $\mathrm{OM}\left( \tilde{Q}_{f_o}^{r} \right)$ & [$F$,$D$] &  & \\ \hline
		\multirow{2}{*}{Eqn.~(7)} & $\mathring{S}_{f_o}^{ck}$ & [$F$,$D$] & $\hat{S}_{f}^{ck}$ & [$F$,$D$] \\
		& $\mathring{Q}_{f_o}^{r}$ & [$F$,$D$] & $\hat{Q}_f^{r}$ & [$F$,$D$] \\ 
		\Xhline{1pt}
	\end{tabular}
	\caption{Model structure of Manta. Major tensor changes are also provided.}
	\label{tab: structure}
\end{table} 
\subsection{Pseudo-Code}
\indent For the analysis of computational complexity. The entire process of Matryoshka Mamba is demonstrated in Algorithm~\ref{alg: MaM}. We find that an outer loop for multi-scale design and a inner loop for local features consist or the main calculation of Matryoshka Mamba. The computational complexity of the outer loop is $O(\left| \mathcal{O} \right|)$ while the inner loop is $O(F)$. From the structure of algorithm, the inner loops are included in an outer loop. Therefore, the computational complexity of the Matryoshka Mamba is $O(F\left| \mathcal{O} \right|)$. From extensive experiments, the two parameters we determined are both small~($F=16$, $\mathcal{O}=\left\{ 1,2,4 \right\}$, $\left| \mathcal{O} \right|=3$). Therefore, the new-designed Matryoshka Mamba does not introduce heavy computational burden into practical use. 
\begin{algorithm}[t]
	\small
	\caption{Matryoshka Mamba}
	\label{alg: MaM}
	\LinesNumbered
	\SetKwFunction{Shape}{Shape}
	\SetKwFunction{Concat}{Concat}
	\SetKwFunction{Sigmoid}{Sigmoid}
	\SetKwFunction{CB}{CB}
	\SetKwFunction{Average}{Average}
	\SetKwFunction{IM}{IM}
	\SetKwFunction{OM}{OM}
	\KwIn{$I=\left[ I_{1},\dots ,I_{F} \right] \in \mathbb{R} ^{F\times D}~\left(F\mid 2\right)$,  $\mathcal{O}~\left(o\in \mathcal{O}, F\mid o, o< F, o= 2^\alpha, \alpha\in \mathbb{Z} ^+\right)$}
	\KwOut{$\hat{I}\left[ \hat{I}_{1},\dots ,\hat{I}_{F} \right]  \mathbb{R} ^{F\times D}~\left(F\mid 2\right)$}
	$\hat{I}\gets \varnothing $\;
	\For{each $o \in \mathcal{O}$}
	{$F \gets \Shape\left(I, 0\right) $; $W \gets F-o+1$\;
		$E \gets \varnothing $; $\tilde{I} \gets \varnothing $; $w \gets \varnothing $; $\mathring{I} \gets \varnothing $\; 
		\For{each $i \in \left[0, W-1\right]$}
		{	\tcc{Inner Module}
			$E \gets \IM\left(I\left[i:i+o, :\right]\right)  \oplus I\left[i:i+o, :\right]$ \;
			\If{$\tilde{I}= \varnothing$}
			{
				$\tilde{I} \gets E$; 
			}
			\Else{
				\tcc{Outer Module}
				$\tilde{I} \gets \OM\left\{\Concat\left[\left(\tilde{I}, E \right), \mathrm{dim}=0\right]\right\} $;
			}
		}
		$w\gets \Sigmoid\left[\CB\left(\tilde{I}\right)\oplus I\right] $\;
		$\mathring{I}\gets w \otimes \tilde{I} $\;
		\If{$\hat{I}= \varnothing$}
		{
			$\hat{I} \gets \mathring{I}$; 
		}
		\Else{$\hat{I} \gets \Average\left(\hat{I}\oplus \mathring{I}\right) $;}
	}
	\Return{$\hat{I}$}
\end{algorithm}
\section{Contribution Statement}
\begin{itemize}
	\item \textbf{Wenbo~Huang~(Southeast University, China):} Proposing the idea, implementing code, conducting experiments, data collection, figure drawing, table organizing, and completing original manuscript.
	\item \textbf{Jinghui~Zhang~(Southeast University, China):} Providing experimental platform, supervision, writing polish, and funding acquisition.
	\item \textbf{Guang~Li~(Hokkaido University, Japan):} Idea improvement, writing polish, rebuttal assistance, and funding acquisition.
	\item \textbf{Lei~Zhang~(Nanjing Normal University, China):} Idea improvement, writing polish, rebuttal assistance, and funding acquisition.
	\item \textbf{Shuoyuan~Wang~(Southern University of Science and Technology, China):} Data verification, writing polish, and rebuttal assistance.
	\item \textbf{Fang~Dong~(Southeast University, China):} Funding acquisition.
	\item \textbf{Jiahui~Jin~(Southeast University, China):} Funding acquisition.
	\item \textbf{Takahiro~Ogawa~(Hokkaido University, Japan):} Idea improvement, writing polish, rebuttal assistance, and funding acquisition.
	\item \textbf{Miki~Haseyama~(Hokkaido University, Japan):} Funding acquisition.
\end{itemize}
\section{Acknowledgements}
The authors would like to appreciate all participants of peer review and cloud servers provided by Paratera Ltd. Wenbo Huang sincerely thanks all family members and TJUPT for the encouragement at an extremely difficult time. This work is supported by Frontier Technologies Research and Development Program of Jiangsu under Grant No.~BF2024070; National Natural Science Foundation of China under Grants Nos.~62472094, 62072099, 62232004, 62373194, 62276063; Jiangsu Provincial Key Laboratory of Network and Information Security under Grant No.~BM2003201; Key Laboratory of Computer Network and Information Integration~(MOE, China) under Grant No.~93K-9; the Fundamental Research Funds for the Central Universities; the Excellent Ph.D Training Program~(SEU); and JSPS KAKENHI Grant Nos.~JP23K21676, JP24K02942, JP24K23849.

\begin{links}
	\link{Code}{https://github.com/wenbohuang1002/Manta}
\end{links}

\bibliography{aaai25}

\end{document}